\pdfoutput=1

\documentclass[11pt]{article}

\usepackage[preprint]{acl}

\usepackage{times}
\usepackage{latexsym}
\usepackage{amssymb}
\usepackage{multirow} 
\usepackage{enumitem}
\usepackage{textcomp}
\usepackage{hyperref}
\usepackage{amsmath}
\usepackage{bm}
\usepackage{subcaption}

\usepackage{colortbl}       

\definecolor{lightblue}{RGB}{173, 216, 230}

\usepackage[T1]{fontenc}

\usepackage[utf8]{inputenc}

\usepackage{microtype}

\usepackage{inconsolata}

\usepackage{graphicx}

\title{Multimodal Fusion and Coherence Modeling for Video Topic Segmentation}

\author{
Hai Yu, Chong Deng, Qinglin Zhang, Jiaqing Liu, Qian Chen, Wen Wang \\
Speech Lab, Alibaba Group \\
\texttt{\{yuhai.yu, w.wang\}@alibaba-inc.com}
}

\begin{document}
\maketitle
\begin{abstract}
The video topic segmentation (VTS) task segments videos into intelligible, non-overlapping topics, facilitating efficient comprehension of video content and quick access to specific content. VTS is also critical to various downstream video understanding tasks. Traditional VTS methods using shallow features or unsupervised approaches struggle to accurately discern the nuances of topical transitions. Recently, supervised approaches have achieved superior performance on video action or scene segmentation over unsupervised approaches. In this work, we improve supervised VTS by thoroughly exploring \textbf{multimodal fusion} and \textbf{multimodal coherence modeling}. Specifically, (1) we enhance multimodal fusion by exploring different architectures using Cross-Attention and Mixture of Experts. (2) To generally strengthen multimodality alignment and fusion, we pre-train and fine-tune the model with multimodal contrastive learning. (3) We propose a new pre-training task tailored for the VTS task, and a novel fine-tuning task for enhancing multimodal coherence modeling for VTS. We evaluate our proposed approaches on \textit{educational videos}, in the form of \textit{lectures}, due to the vital role of topic segmentation of educational videos in boosting learning experiences. Additionally, to promote research in VTS, we introduce a large-scale Chinese lecture video dataset to augment the existing English lecture video datasets. Experiments on both English and Chinese lecture datasets demonstrate that our model achieves superior VTS performance compared to competitive unsupervised and supervised baselines\footnote{The code and model checkpoints will be released upon acceptance}.
\end{abstract}

\section{Introduction}
The proliferation of digital video content over the last few decades has underscored the importance of efficient content navigation and comprehension. As the unstructured nature of videos poses significant challenges for users seeking to quickly grasp or reference specific topics, Video Topic Segmentation (VTS) has emerged as a vital tool in addressing these demands. By delineating videos into coherent non-overlapping topics, VTS not only facilitates intuitive understanding of video content but also enables swiftly pinpointing and accessing topics of interest. This is particularly pertinent for the furtherance of various video understanding tasks, where VTS serves as a foundational component.

Traditional VTS approaches predominantly hinge on shallow features~\cite{gandhi2015topic, soares2018framework, ali2021segmenting} and unsupervised methods~\cite{gupta2023unsupervised}, due to scarcity of labeled data. These methods often fall short in capturing the semantic cues that signal topical shifts in video streams, hence suffer from limited precision. Recent advancements in supervised learning paradigms have achieved notable performance improvements in multi-modal task~\cite{yang2022paraphrasing,tu20222,tu2023self,zhang2022mcse} and various video segmentation tasks, such as video action segmentation~\cite{zhou2018towards, tang2019coin}, scene segmentation~\cite{huang2020movienet, islam2023efficient}, and topic segmentation~\cite{wu2023newsnet, wang2023vstar, xing2024multi}, surpassing unsupervised methods. Performance of supervised approaches can be further enhanced by pre-training on vast volumes of unlabeled data~\cite{xu2021videoclip, mun2022bassl} or initializing models from pre-trained models~\cite{yan2023unloc} and then fine-tuning the model. Hence, in this work, we focus on further improving supervised methods for VTS.

Compared to text topic segmentation~\cite{koshorek2018text,xing2021improving, yu2023improving}, videos contain rich and diverse multimodal contextual information. Fully utilizing multimodal information, such as visual cues and textual data (e.g., screen text and subtitles), could facilitate more detailed content understanding and in turn more accurate semantic segmentation than relying on text only. Our case studies in Appendix~\ref{appendix_qualitative_analysis} demonstrate the great challenges posed by VTS, particularly to unsupervised approaches or supervised methods that rely solely on either visual or textual modality. The complexity inherent in video content—where multimodal signals must be effectively integrated—accentuates the difficulty.  Also, coherence is essential for understanding logical structures and semantics. Enhancing coherence modeling has achieved significant improvements in long text topic segmentation~\cite{yu2023improving}. Therefore, we improve supervised VTS methods by thoroughly exploring \textbf{multimodal fusion} and \textbf{multimodal coherence modeling}. We enhance multimodal fusion from the perspectives of model architecture and pre-training and fine-tuning tasks. Specifically, we compare various multimodal fusion architectures built upon Cross-Attention and Mixture-of-Experts (MoE). We investigate the effect of multimodal contrastive learning for general pre-training and fine-tuning for strengthening cross-modal alignment. For enhancing multimodal coherence modeling, we propose a new pre-training task tailored for the VTS task, and a novel fine-tuning task by elevating intra-topic multimodal feature similarities and inter-topic multimodal feature differences. The proposed approaches are extensively evaluated on educational videos, in the form of lectures, due to the pivotal contributions of topic segmentation of educational videos in bolstering the learning experiences.

Our contributions can be summarized as follows.
\begin{itemize}[leftmargin=*,noitemsep]
    \item We propose a supervised multimodal sequence labeling model for VTS, denoted \textbf{MMVTS}. We explore various multimodal fusion architectures, and apply multimodal contrastive learning for strengthening cross-modal alignment. We also propose a new self-supervised pre-training task tailored to the VTS and a novel fine-tuning task for enhancing multimodal coherence modeling.
    \item We introduce a large-scale \textbf{C}hinese \textbf{L}ecture \textbf{V}ideo \textbf{T}opic \textbf{S}egmentation dataset (\textbf{CLVTS}) to promote the research of VTS.
    \item Experiments show that our model sets new state-of-the-art (SOTA) performance on both English and Chinese lecture video datasets, outperforming competitive unsupervised and supervised baselines. Comprehensive ablation study further confirms the effectiveness of our approaches.
\end{itemize}

\begin{table*}[]
    \centering
    \resizebox{0.9\textwidth}{!}{
    \begin{tabular}{llcccclll}
        \hline
        \textbf{Dataset} & \textbf{Videos} & \textbf{Hours} & \textbf{Topics/Video} & \textbf{Clips/Topic} & \textbf{Seconds/Clip} & \textbf{Domain} & \textbf{Language} & \textbf{Available} \\
        \hline
        NPTEL10~\cite{gandhi2015topic} & 12 & - & - & - & - & Education & English & \(\times\) \\
        Videoaula~\cite{soares2018automatic} & 44 & 26.4 & - & - & - & Education & Portuguese & \checkmark \\
        CS80~\cite{soares2019optimization} & 80 & - & - & - & - & Education & English & \checkmark \\
        MOOC100~\cite{das2019automatic} & 100 & 100 & 6.9 & - & - & Education & English & \checkmark$\dagger$ \\
        Coursera37~\cite{chand2021framework} & 37 & 2.8 & 16.5 & - & - & Education & English & \(\times\) \\
        VSTAR~\cite{wang2023vstar} & 8159 & 4625 & 61.2 & 0.4 & 90 & Television & English & \checkmark$\dagger$ \\
        NewsNet~\cite{wu2023newsnet} & 1000 & 946 & 8.5 & - & - & News & English & \(\times\) \\
        MultiLive~\cite{qiu2023liveseg} & 1000 & 1300 & 8.8 & - & - & Livestream & English & \(\times\) \\
        \rowcolor{lightblue} AVLecture~\cite{gupta2023unsupervised} & 350 & 297.5 & 5.4 & 46.2 & 12.3 & Education & English & \checkmark \\
        YouTube~\cite{xing2024multi} & 5422 & 858.5 & 6.7 & 16 & 5.3 & Diverse & English & \(\times\) \\
        Behance~\cite{xing2024multi} & 575 & 1225.2 & 5.2 & 248 & 6.0 & Livestream & English & \(\times\) \\
        \rowcolor{lightblue} \textbf{CLVTS (Ours)} & 510 & 395 & 10.1 & 35.7 & 7.7 & Education & Chinese & \checkmark \\
        \hline
    \end{tabular}}
    \caption{Comparison between our \textbf{CLVTS} dataset and existing video datasets for the video topic segmentation task. $\dagger$ indicates that the data is not entirely open source. Prior to our work, AVLecture is the only publicly available large-scale video dataset supporting supervised VTS methods.}
    \label{table_data_stats}
\end{table*}

\section{Related Work}
\noindent \textbf{Text Topic Segmentation}
Text topic segmentation aims to automatically partition text into topically consistent, non-overlapping segments~\cite{hearst1994multi}. By automatically mining clues of topic shifts from large amounts of labeled data~\cite{koshorek2018text, arnold2019sector}, contemporary supervised models~\cite{lukasik2020text, somasundaran2020two, zhang2021sequence, yu2023improving} demonstrate superior performance compared to unsupervised approaches~\cite{riedl2012topictiling, solbiati2021unsupervised}. Notably, supervised models that excel at modeling long sequences~\cite{zhang2021sequence, yu2023improving} are capable of capturing longer contextual nuances and thereby achieve better topic segmentation performance, compared to models that model local sentence pairs or block pairs~\cite{wang2017learning, lukasik2020text}. In addition, recent works~\cite{somasundaran2020two, xing2020improving, yu2023improving} show that strengthening coherence modeling can improve text topic segmentation performance. Inspired by these findings, in this work, we explore enhancing coherence modeling for video topic segmentation under the \textit{multimodal} configurations.

\noindent \textbf{Video Topic Segmentation}
For video topic segmentation, some approaches, such as BaSSL~\citet{mun2022bassl}, explore visual-only information. However, many recent works have achieved enhanced semantic understanding of videos by leveraging multimodal data. \citet{gupta2023unsupervised} introduced UnsupAVLS, which uses the TWFINCH algorithm to cluster video clips into topics based on visual and text features. \citet{wang2023vstar} proposed SWST, which concatenates visual and text features for language models; however, it may suffer from discrepancies between pre-training of the language model and fine-tuning. \citet{wu2023newsnet} focused on hierarchical modeling of scene, story, and topic, without further exploring how to better integrate multimodal features. \citet{xing2024multi} employed asymmetric cross-modal attention for obtaining text-aware visual representations. It may be most related to our work. However, our work differs from \citet{xing2024multi} as we explore symmetric cross-modal attention and also investigate the Mixture-of-Experts mechanism, as well as introducing topic-level Contrastive Semantic Similarity Learning into fine-tuning for enhanced coherence modeling in the multimodal framework.

\section{Methodology}
\begin{figure*}[htbp]
    \centering
    \begin{minipage}[c]{0.48\textwidth}
        \includegraphics[width=\linewidth]{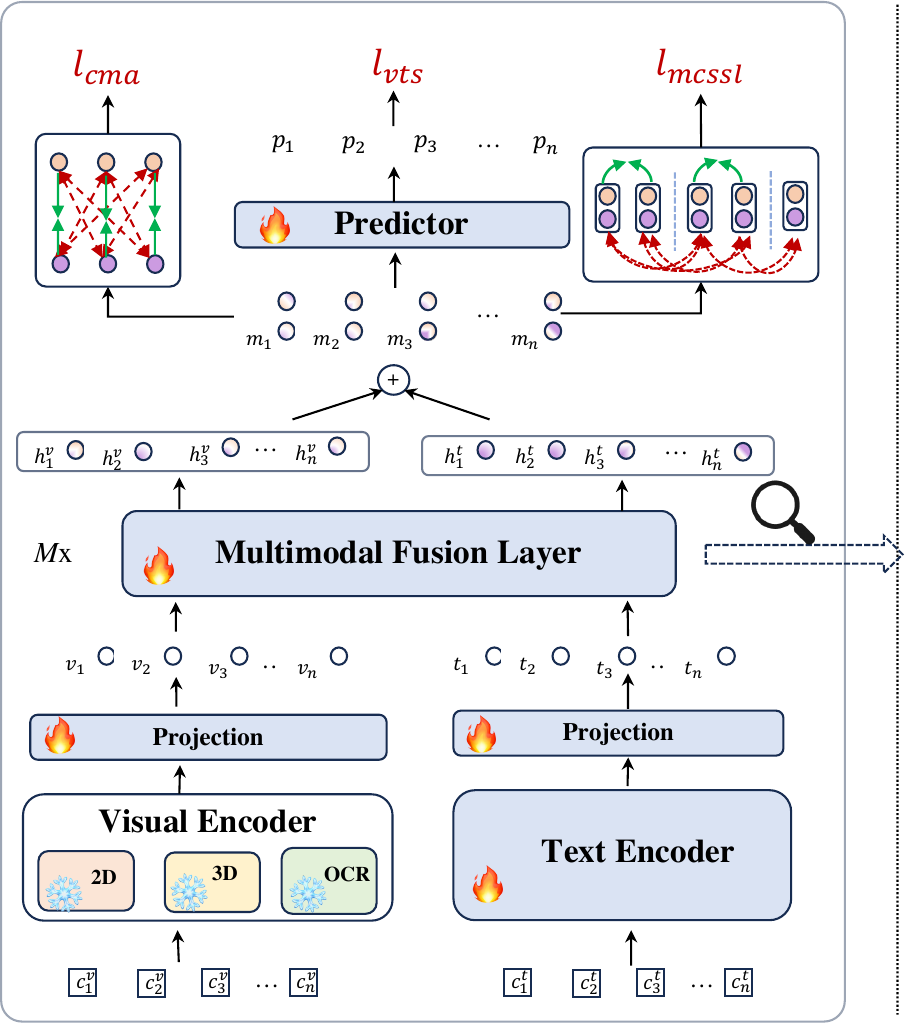}
        \subcaption{Overall Model Architecture}
        \label{figure_overall_model}
    \end{minipage}
    \hfill
    \begin{minipage}[c]{0.5\textwidth}
        \begin{minipage}{\textwidth}
            \begin{minipage}{0.48\textwidth}
                \includegraphics[width=\linewidth]{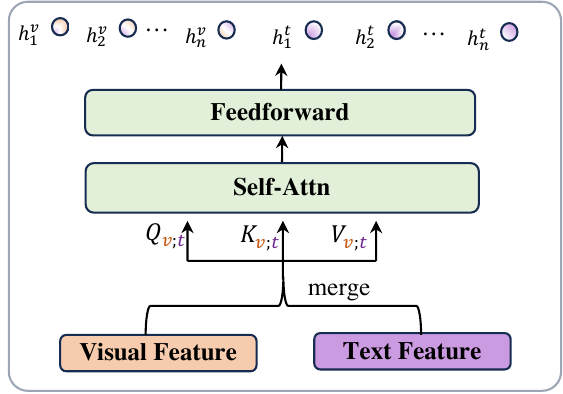}
                \subcaption{Merge-Attention}
                \label{figure_merge_attn}
            \end{minipage}
            \hfill
            \begin{minipage}{0.48\textwidth}
                \includegraphics[width=\linewidth]{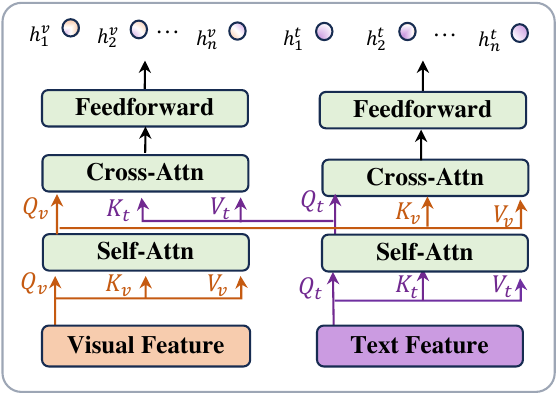}
                \subcaption{Co-Attention}
                \label{figure_co_attn}
            \end{minipage}
        \end{minipage}
        \vspace{5pt}
        \begin{minipage}{\textwidth}
            \begin{minipage}{0.48\textwidth}
                \includegraphics[width=\linewidth]{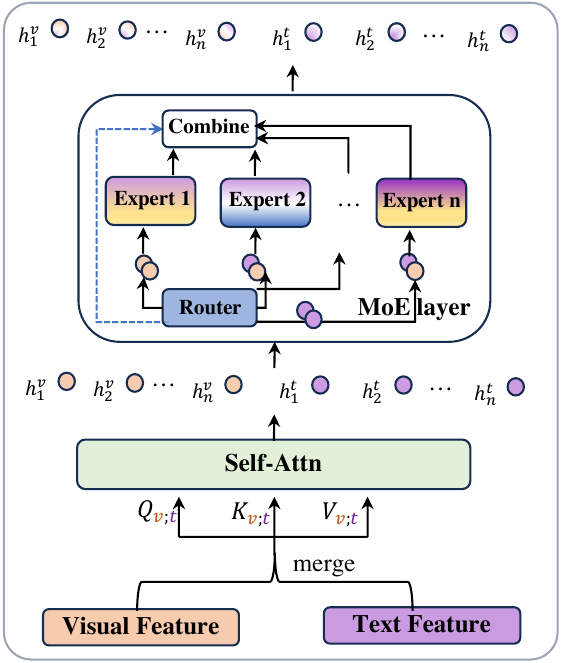}
                \subcaption{Merge-Attention with MoE}
                \label{figure_merge_attn_moe}
            \end{minipage}
            \hfill
            \begin{minipage}{0.48\textwidth}
                \includegraphics[width=\linewidth]{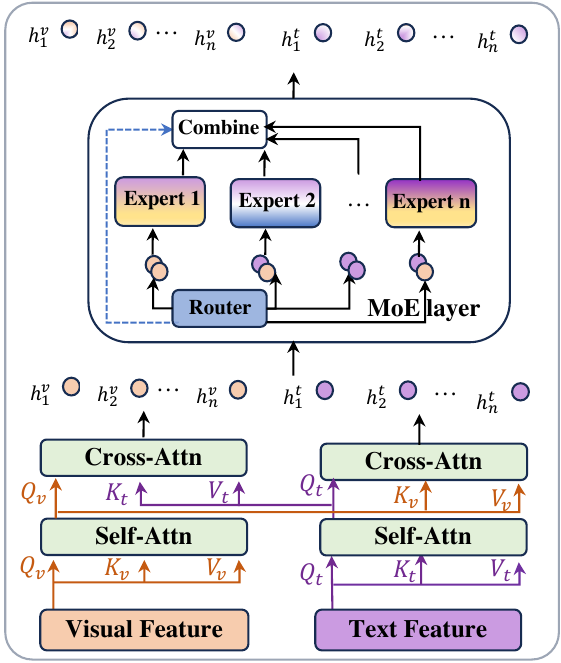}
                \subcaption{Co-Attention with MoE}
                \label{figure_co_attn_moe}
            \end{minipage}
        \end{minipage}
    \end{minipage}

    \caption{The overall architecture of our \textbf{MMVTS model} and four distinct architectures of the Multimodal Fusion Layers in (a). In the overall architecture, the snowflake symbol indicates that the parameters of the certain module are frozen; whereas, the flame symbol signifies a trainable module. The blue dotted lines in the $l_{mcssl}$ module denote the topic boundaries. The green solid lines in the $l_{cma}$ module depict the features being brought closer, while the red dashed lines depict the features being pushed apart.}
    \label{figure_model}
\end{figure*}

Figure~\ref{figure_model} depicts the overall architecture of our MMVTS model. Section~\ref{subsec:mm-vts} presents the problem definition of multimodal VTS and the overall model architecture. We enhance multimodal fusion from the perspectives of model architecture, pre-training, and fine-tuning tasks. Specifically, we compare different multimodal fusion architectures built upon Merge- and Cross-Attention, and Mixture-of-Experts (Section~\ref{subsec:mm-vts}). We explore multimodal contrastive learning for cross-modality alignment and propose a new pre-training task tailored for VTS (Section~\ref{subsec:pretraining}). For fine-tuning, we also propose a novel task for multimodal coherence modeling (Section~\ref{subsec:finetuning}).

\subsection{MultiModal Video Topic Segmentation}
\label{subsec:mm-vts}
\paragraph{\textbf{Overall Architecture.}} Following prior works~\cite{zhang2021sequence, wu2023newsnet}, we define video topic segmentation as a clip-level sequence labeling task and propose our \textbf{M}ulti\textbf{M}odal \textbf{V}ideo \textbf{T}opic \textbf{S}egmentation (MMVTS) model. As illustrated in Figure~\ref{figure_overall_model}, we apply unimodal pre-trained encoders for the vision and text modality, respectively, and then fuse multimodal information at the intermediate representation level (i.e., \textit{middle fusion}~\cite{xu2023multimodal}) through the \textbf{Multimodal Fusion Layer}. Given a video, we transcribe it with a competitive automatic speech recognition (ASR) system\footnote{\url{https://tingwu.aliyun.com/home}} and use ASR 1-best as the text modality. We then divide the video into $n$ clips $(c^{v}_{i}, c^{t}_{i})_{i=1}^{n}$, with clips segmented at the sentence boundaries predicted on ASR 1-best. $c^{v}_{i} = \{f^{i}_{1},..., f^{i}_{k}\}$ denotes evenly sampled $k$ frames within the $i$-th clip and is fed into a visual encoder $\bm{E_{v}}$ to extract visual features. $c^{t}_{i} = \{w^{i}_{1},...,w^{i}_{\|s_i\|+1}\}$ denotes the sequence of words from ASR 1-best within the $i$-th clip, where $w^{i}_1$ is the inserted special token [BOS] and $\|s_i\|$ denotes the number of words in the $i$-th clip. $c^{t}_{i}$ is fed into a text encoder $\bm{E_{t}}$ and the last hidden representation of [BOS] for the i-th clip is used as the text representation of the clip. After extracting the unimodal features, we first apply trainable projection matrices $W_{v}$ and $W_{t}$ to convert unimodal features into the same dimension, resulting in the visual feature sequence $\bm{v}=\{v_{1},...,v_{n}\}$ and the textual feature sequence $\bm{t}=\{t_{1},...,t_{n}\}$ (Eq.~\ref{formula_encoder_projector}). Then we fuse the multimodal information with $M$ Multimodal Fusion Layers $\bm{MFL}_{M}$ and obtain the updated visual features $\bm{h^{v}}=\{h^{v}_{1},...,h^{v}_{n}\}$ and textual features $\bm{h^{t}}=\{h^{t}_{1},...,h^{t}_{n}\}$ (Eq.~\ref{formula_modality_fusion_layer}), which are then concatenated into the multimodal features $\bm{m}=\{m_{1},...,m_{n}\}$ (Eq.~\ref{formula_cat}). Finally, the multimodal features $\bm{m}$ are fed into the predictor consisting of a linear layer $W_{p}$ to obtain the probability of binary classification $\bm{p}=\{p_{1},...,p_{n}\}$ (Eq.~\ref{formula_predictor}), where $p_{i}$ indicates whether the $i$-th clip is at a topic boundary. We use the standard binary cross-entropy loss (Eq.~\ref{formula_l_vts}) to train the model, where $y_{i} \in \{0, 1\}$ is the label. The last clip is excluded from loss computation. \textbf{Considering computational complexity, we freeze the visual encoder while keeping all other parameters trainable}.

Compared to \textit{late fusion} where no cross-modal interaction happens until after independent predictions by each unimodal model, \textit{middle fusion} and \textit{early fusion} are found to generally outperform late fusion~\cite{nagrani2021attention}, probably because early and middle fusion aligns better with human perception where multimodal fusion happens early in sensory processing. On the other hand, compared to early fusion, middle fusion yields superior or comparable performance~\cite{nagrani2021attention} and is much less computationally expensive since we could freeze some strong pre-trained unimodal encoders with a large number of parameters and only train a small number of parameters.

\begin{equation}
\small
\begin{aligned}
    \bm{v} = W_{v} \cdot \bm{E_{v}}(\{c^{v}_{1},...,c^{v}_{n}\})\\
    \bm{t} = W_{t} \cdot \bm{E_{t}}(\{c^{t}_{1},...,c^{t}_{n}\})
\end{aligned}
    \label{formula_encoder_projector}
\end{equation}
\begin{equation}
\small
    \bm{h^{v}}; \bm{h^{t}} = \bm{MFL}_{M}(\bm{v};\bm{t})
    \label{formula_modality_fusion_layer}
\end{equation}
\begin{equation}
\small
    m_{i} = h^{v}_{i};h^{v}_{i}
    \label{formula_cat}
\end{equation}
\begin{equation}
\small
    \bm{p} = W_{p} \cdot \bm{m}
    \label{formula_predictor}
\end{equation}
\begin{equation}
    \small
    l_{vts} = -\sum_{i=1}^{n-1}[y_{i}\ln p_{i}+(1\!-\!y_{i})\ln(1\!-\!p_{i})]
    \label{formula_l_vts}
\end{equation}

\paragraph{\textbf{Multimodal Fusion Layer (MFL).}} We compare four distinct cross-modal interaction mechanisms for Multimodal Fusion Layers. We investigate the Merge-Attention and Co-Attention multimodal fusion layers proposed in~\cite{dou2022empirical, yang2023code}. With \textbf{\textit{Merge-Attention}} (Figure~\ref{figure_merge_attn}), features from unimodal encoders are concatenated sequentially and then input into a standard transformer encoder layer~\cite{vaswani2017attention}, which \textit{shares} attention parameters across modalities. A feed forward layer is added on top to produce the final output representation. In contrast, with \textbf{\textit{Co-Attention}} (Figure~\ref{figure_co_attn}),  features from each unimodal encoder first go through self-attention with \textit{modality-specific} attention parameters, then we perform \textbf{symmetric cross-attention} to integrate information from all other modalities to enhance the representation of the considered modality, followed by a feed forward layer.

Inspired by~\cite{mustafa2022multimodal}, which interleaves MoE encoder layers and standard dense encoder layers for image-text multimodal models, we also investigate two new architectures by replacing the traditional single feed-forward layers in Figure~\ref{figure_merge_attn} and \ref{figure_co_attn} with a MoE module~\cite{shazeer2016outrageously, lepikhin2020gshard}. The resulting architectures are depicted by Figure~\ref{figure_merge_attn_moe} and \ref{figure_co_attn_moe}. The motivation is that adding MoE on top of the fused representations may \textit{facilitate deeper cross-modal integration of information} and \textit{improve model capacity without a proportional increase in computational complexity}. Specifically, \textit{experts} are MLPs activated depending on the input. Firstly, we concatenate fused features output from self-attention or cross-attention. Then, we implement the Noisy Top-k Gating mechanism~\cite{shazeer2016outrageously} to select $K$ experts from a total of $E$ candidates (Eq.~\ref{formula_moe_router} -~\ref{formula_moe_topk}), where $SN()$ denotes the standard normal distribution, $W_{n}$ denotes tunable Gaussian noise to help load balancing, $W_g$ is a trainable weight matrix, $K$ and $E$ are hyper-parameters. Finally, the outputs of the $K$ activated experts are linearly combined with the learned gating weights  (Eq.~\ref{formula_moe_combine}). For the MoE training objective, we sum the \textit{importance} loss and the \textit{load} loss~\cite{shazeer2016outrageously} to balance expert utilization as in Eq.~\ref{formula_moe_balance}.

\begin{equation}
\small
    G(x) = Softmax(KeepTopK(H_{x}, k))
    \label{formula_moe_router}
\end{equation}
\begin{equation}
\small
    H(x)_{i} = (W_{g} \cdot x)_{i} + SN() \cdot Softplus((W_{n} \cdot x)_{i})
\end{equation}
\begin{equation}
\small
    KeepTopK(x, k)_{i} =
    \begin{cases}
        x_{i}~\text{if}~x_{i}~\text{is in top-k.}\\
        -\infty~\text{otherwise.}
    \end{cases}
    \label{formula_moe_topk}
\end{equation}
\begin{equation}
\small
    MoE(x) = \sum^{K}_{e=1} G(x)_{e} \cdot MLP_{e}(x)
    \label{formula_moe_combine}
\end{equation}
\begin{equation}
\small
    l_{balance} = l_{importance} + l_{load}
    \label{formula_moe_balance}
\end{equation}

\begin{table*}[]
    \centering
    \resizebox{0.96\textwidth}{!}{
    \begin{tabular}{l|c|l|c|ccccc|ccccc}
    \hline
    \multicolumn{3}{l|}{\textbf{Model}} & \textbf{Modality} & \multicolumn{5}{|c|}{\textbf{AVLecture}} & \multicolumn{5}{|c}{\textbf{CLVTS}} \\
    \multicolumn{3}{l|}{} & \multicolumn{1}{|l|}{} & \textbf{\textit{F$_{1}$}} & \textbf{\textit{BS@30}} & \textbf{\textit{F$_{1}$@30}} & \textbf{\textit{mIoU}} & \textbf{\textit{Avg}} & \textbf{\textit{F$_{1}$}} & \textbf{\textit{BS@30}} & \textbf{\textit{F$_{1}$@30}} & \textbf{\textit{mIoU}} & \textbf{\textit{Avg}} \\
    \hline
    \multicolumn{3}{l|}{UnsupAVLS~\cite{gupta2023unsupervised}} & V+T & - & 56.00$\ddagger$ & - & \underline{70.86}$\ddagger$ & - & - & - & - & - & -\\ 
    \multicolumn{3}{l|}{BaSSL~\cite{mun2022bassl}} & V & - & 43.94 & - & 46.95 & - & - & - & - & - & -\\
    \multicolumn{3}{l|}{LongFormer~\cite{yu2023improving}} & T & 52.91 & 69.25 & 60.38 & 67.54 & 62.52 & 34.42 & 52.19 & 47.77 & 52.87 & 46.81\\
    \multicolumn{3}{l|}{LongFormer$_{cssl}$~\cite{yu2023improving}} & T & \underline{54.02} & \underline{71.56} & 62.40 & 68.39 & \underline{64.09} & \underline{34.77} & 53.07 & 47.51 & 53.15 & 47.12\\
    \multicolumn{3}{l|}{Llama-3-8B$_{Generative}$} & T & 40.00 & 57.55 & 56.52 & 62.8 & 54.22 & 27.50 & 40.58 & 43.71 & 50.38 & 40.54\\
    \multicolumn{3}{l|}{Llama-3-8B$_{Discrete}$} & T & 39.27 & 68.8 & \underline{62.55} & 70.43 & 60.26 & 31.47 & \underline{60.40} & \textbf{54.64} & \textbf{58.86} & \textbf{51.34}\\
    \multicolumn{3}{l|}{SWST$_{seq}$~\cite{wang2023vstar}} & V+T & 53.45 & 70.95 & 59.73 & 65.21 & 62.33 & 34.55 & 52.77 & 48.08 & 52.67 & 47.02\\
    \hline
    \multicolumn{1}{c|}{\textbf{PT}} & \multicolumn{1}{|c|}{\textbf{FT-Coh}} & \textbf{MMVTS Models (Ours)} & \textbf{Modality} &\\
    \hline
    \texttimes & \texttimes & Baseline$_{1}$ & \multirow{3}{*}{V+T} & 55.19 & 71.76 & 61.19 & 66.39 & 63.63 & \underline{37.32} & 49.75 & 47.07 & 50.51 & 46.16\\
    \texttimes & \checkmark & Baseline$_{2}$ & & 56.72 & \underline{72.56} & 63.03 & 67.97 & 65.07 & 37.29 & 48.48 & 47.62 & 51.73 & 46.28\\
    \checkmark & \checkmark & Baseline$_{3}$ & & \underline{58.77} & 72.55 & \underline{67.26} & \underline{71.52} & \underline{67.52} & 36.54 & \underline{50.67} & \underline{48.81} & \underline{52.56} & \underline{47.15}\\
    \hline
    \checkmark & \checkmark & Merge-Attn & \multirow{4}{*}{V+T} & 57.36 & 74.96 & 65.30 & 70.15 & 66.94 & 38.17 & 55.52 & 50.69 & 54.84 & 49.80\\
    \checkmark & \checkmark & Co-Attn & & \textbf{60.01} & 73.88 & 67.27 & \textbf{72.32} & 68.37 & 38.49 & 57.23 & 50.59 & 54.47 & 50.20\\
    \checkmark & \checkmark & Merge-Attn with MoE & & 57.54 & 73.48 & 64.36 & 70.43 & 66.45 & 38.77 & \textbf{61.05} & 51.10 & 54.41 & \underline{51.33}\\
    \checkmark & \checkmark & Co-Attn with MoE & & 59.77 & \textbf{75.01} & \textbf{67.94} & 71.69 & \textbf{68.61} & \textbf{39.98} & 58.96 & \underline{51.41} & \underline{54.71} & 51.27\\
    \hline
    \end{tabular}}
    \caption{Performance of baselines and our MMVTS models on AVLecture and CLVTS test sets. $\ddagger$ denotes the leakage of the ground-truth topic number.  \textbf{V} and \textbf{T} under \textbf{Modality} denote Vision and Text modality, respectively. MMVTS Baseline$_{1,2,3}$ denote our MMVTS model w/o Multimodal Fusion Layers. Attn denotes Attention. \textbf{PT} denotes pre-training the model on unlabeled data (Section~\ref{subsec:pretraining} Eq.~\ref{formula_l_pre}) before fine-tuning. \textbf{FT-Coh} denotes adding the two auxiliary multimodal coherence modeling tasks during fine-tuning (Section~\ref{subsec:finetuning} Eq.~\ref{formula_l_finetune}); w/o FT-Coh refers to fine-tuning with the standard $l_{vts}$ (Eq.~\ref{formula_l_vts}). For each metric, the best result among all models is boldfaced while the best result in each group is underscored.}
    \label{table_main_results}
\end{table*}

\subsection{Pre-training with Unlabeled Data}
\label{subsec:pretraining}
Prior works have demonstrated that standard self-supervised denoising pre-training (even only using the downstream task data)~\cite{amos2023never} or pre-training adapted to the downstream task~\cite{gururangan2020don} often perform substantially better than randomly initializing the parameters. Therefore, to better initialize the parameters of the Multimodal Fusion Layers, we explore pre-training with unlabeled video data before supervised fine-tuning. Firstly, we introduce \textbf{a general cross-modality alignment pre-training task} to learn the multimodal representation. We use contrastive learning loss to adjust the features learned by the Multimodal Fusion Layers, by maximizing the cosine similarity of the visual features and textual features of the same clip, while reducing the similarity of the modality features between different clips, as show in Eq.~\ref{formula_l_cma}, where $\epsilon$ is used to prevent division by 0 and $\tau$ is a temperature hyper-parameter to scale the cosine similarity.

\begin{equation}
\small
    l_{cma} = -\frac{1}{n} \frac{\sum_{i=1}^{n} e^{sim(h^{v}_{i}, h^{t}_{i})}}{\sum_{i=1}^{n}\sum_{j=1}^{n} e^{sim(h^{v}_{i}, h^{t}_{j})}+\epsilon}\\
    \label{formula_l_cma}
\end{equation}
\begin{equation}
\small
    sim(x_{1}, x_{2}) = \frac{x_{1}^{\mathrm{T}} \cdot x_{2}}{\left\|x_{1}\right\| \cdot \left\|x_{2}\right\|} / \tau
    \label{formula_sim}
\end{equation}
Secondly, we introduce \textbf{a novel pre-training task tailored for the VTS task}, focusing on utilizing unlabeled data for learning pseudo topic boundaries and also enhancing modality alignment. We apply a Kernel Density Estimation (KDE)~\cite{davis2011remarks} model to estimate the topic duration distribution within the labeled training set. Videos are segmented based on KDE-sampled durations. For each segment, with equal probability, we: insert a random segment from other videos, replace it with another, or retain it. These modified segments serve as distinct topics, allowing the model to learn pseudo topic boundaries during pre-training. This task-adaptive pre-training task has the same $l_{vts}$ objective as shown in Eq.~\ref{formula_l_vts}. The overall pre-training objective is shown in Eq.~\ref{formula_l_pre}, where $\alpha$ and $\beta$ are hyper-parameters to adjust the loss weights.
\begin{equation}
\small
    l_{pretrain} = l_{vts} + \alpha l_{cma} + \beta l_{balance}
    \label{formula_l_pre}
\end{equation}

\subsection{Fine-tuning with Multimodal Coherence Modeling}
\label{subsec:finetuning}
For fine-tuning, we introduce two auxiliary tasks to enhance multimodal coherence modeling. The cross modal alignment task is the same as the task in Eq.~\ref{formula_l_cma} used in pre-training. This continuity ensures that the modalities retain their coherence through both pre-training and fine-tuning stages, fostering a consistent interplay between different modalities. 
In addition, we adapt the Contrastive Semantic Similarity Learning (CSSL) task proposed by~\citet{yu2023improving}, which leverages the inherent characteristics of topic-related coherence, to the multimodal context. We adopt the same strategy for selecting positive and negative sample pairs~\cite{yu2023improving}, but extend the features to the multimodal representations, as shown in Eq.~\ref{formula_l_mcssl}, where $k_{1}$ and $k_{2}$ are hyper-parameters that determine the number of positive and negative pairs. For each clip's multimodal representation $m_{i}$,  $m_{i,j}^{+}$ denotes the multimodal representation of the $j$-th similar clip in the same topic as clip $i$, while $m_{i,j}^{-}$ denotes the multimodal representation of the $j$-th dissimilar clip in a different topic from clip $i$. We hypothesize that this extension could improve multimodal representation learning by identifying relative consistency relations within topics and across topics.
\begin{equation}
\small
l_{mcssl}=-\frac{1}{n}\sum_{i=1}^{n}log \frac{\sum\limits_{j=1}^{k_{1}} e^{sim(m_{i}, m_{i,j}^{+})} }{\sum\limits_{j=1}^{k_{1}} e^{sim(m_{i}, m_{i,j}^{+})}\!+\!\sum\limits_{j=1}^{k_{2}} e^{sim(m_{i}, m_{i,j}^{-})}}
    \label{formula_l_mcssl}
\end{equation}
The overall fine-tuning objective combines Eq.~\ref{formula_l_vts}, ~\ref{formula_moe_balance}, ~\ref{formula_l_cma}, and ~\ref{formula_l_mcssl},  as shown in Eq.~\ref{formula_l_finetune}, where $\sigma$, $\theta$, and $\gamma$ are hyper-parameters to adjust loss contribution. When the Multimodl Fusion Layers do not contain MoE structure, $\beta$ in Eq.~\ref{formula_l_pre} and $\sigma$ are set to zero.
\begin{equation}
    \small
    l_{finetune}=l_{vts} + \sigma l_{balance} + \theta l_{mcssl} + \gamma l_{cma}
    \label{formula_l_finetune}
\end{equation}

\vspace{-5mm}
\section{Experiments}
\vspace{-2mm}
\subsection{Experimental Setup}
\noindent \textbf{Datasets.} Table~\ref{table_data_stats} summarizes the statistics of various VTS datasets. It clearly shows that \textit{prior to our work, AVLecture~\cite{gupta2023unsupervised} is the only publicly available large-scale labeled video dataset facilitating supervised VTS methods}. To promote the research in VTS, we introduce a large-scale labeled Chinese Lecture Video Topic Segmentation dataset (\textbf{CLVTS}). 
Both AVLecture and CLVTS are sourced from educational videos, where VTS significantly enhances learning experiences. In terms of differences, \textbf{\textit{in addition to the linguistic distinctness from the English lecture dataset AVLecture, CLVTS is characterized by its natural and uninterrupted long videos, a stark contrast to AVLecture}}, since nearly two-thirds of AVLecture are reassembled pre-segmented short videos. As shown in Table~\ref{table_data_stats}, CLVTS features a higher average number of topics per video than AVLecture. Details of the data collection and annotation procedure and analysis of the CLVTS dataset are in Appendix~\ref{subsec:dataset}. \textbf{\textit{Importantly, we put careful ethical considerations for the datasets used in this research in Appendix~\ref{appendix_ethical_considerations}}}.

\noindent \textbf{Baselines and Implementation Details.}
The implementation details are in Appendix~\ref{appendix_training_details}. We carefully select the following representative baselines.

\noindent \textbf{- UnsupAVLS}~\cite{gupta2023unsupervised} is an unsupervised approach that clusters video clips into a pre-defined number of topics, based on visual and text embeddings learned from matching the narration with the temporally aligned visual content.

\noindent \textbf{- Visual-only BaSSL}~\cite{mun2022bassl} is initially proposed for \textit{video scene segmentation}. We use their released checkpoints to initialize our model and fine-tune on the VTS task to evaluate the performance of a \textit{visual-only} model. 

\noindent \textbf{- Text-only LongFormer} is evaluated on long document topic segmentation by~\citet{yu2023improving}. We fine-tune LongFormer and LongFormer$_{cssl}$ in~\cite{yu2023improving} to evaluate the performance of a \textit{text-only} model w/o and w/ Contrastive Semantic Similarity Learning (CSSL) on the VTS task.

\noindent \textbf{- Llama-3-8B} Our pre-training (Section~\ref{subsec:pretraining}) is conducted on the relatively limited unlabeled videos of AVLectures and CLVTS datasets. To investigate the effect of fine-tuning a powerful pre-trained text large language model (LLM) on VTS, we fine-tune Llama-3-8B\footnote{\url{https://huggingface.co/meta-llama/Meta-Llama-3-8B}} with 8B parameters, using two different prompts (see Appendix~\ref{appendix_llama} for details).

\noindent \textbf{- SWST$_{seq}$} is our adapted version of the multimodal video scene and topic segmentation model~\cite{wang2023vstar} with a pre-trained LongFormer~\cite{beltagy2020longformer} as the backbone to VTS, for comparing performance between their \textit{early fusion} and our \textit{middle fusion} strategy on VTS.

In this work,  we choose not to utilize pre-trained vision-language models such as~\cite{yang2023vid2seq, nguyen2024video} due to their limitations in processing long video content as the case of educational videos, although these models demonstrate strong performance on short video clips lasting several seconds. In future work, we plan to enhance the long video understanding capabilities of large multimodal models~\cite{zou2024seconds, zhou2024mlvu} to enable their applications to VTS.

\noindent \textbf{Evaluation Metrics.} We adopt four commonly used metrics, including positive \textit{F$_{1}$}~\cite{zhang2021sequence} (denoted as \textbf{\textit{F$_{1}$}} for brevity), \textbf{BS@k}~\cite{gupta2023unsupervised}, \textbf{\textit{mIoU}}~\cite{mun2022bassl}, and \textbf{\textit{F$_{1}$@k}}. Definitions of the four metrics are in Appendix~\ref{appendix_metrics}. Following \citet{gupta2023unsupervised}, we set k to 30 seconds. We compute the average of these four metrics, denoted by \textbf{\textit{Avg}}, to measure the overall performance of a model.

\vspace{-2mm}
\subsection{Results and Analysis}
\vspace{-2mm}
Table~\ref{table_main_results} compares the performance of baselines (the first group) and variants of our MMVTS models (the second and the third group).

\noindent \textbf{Unimodal performance.} For BS@30 on AVLecture, the text-only Longformer (Row 3) outperforms the visual-only BaSSL (Row 2) by a large gain (+25.31), and also surpasses the unsupervised UnsupAVLS by a notable gain (+13.25). Such results are expected since \textbf{\textit{the text modality inherently conveys more precise information for VTS than the vision modality}}. Notably, the high \textit{mIoU} of the unsupervised method is attributable to the leakage of the ground-truth number of topics.

\noindent \textbf{Mutimodal performance.} As shown in Table~\ref{table_main_results}, \textbf{\textit{Avg}} (the average of \textit{F$_{1}$}, \textit{BS@k}, \textit{mIoU}, \textit{F$_{1}$@k}) of the multimodal model $SWST_{seq}$ is only comparable to Avg of the text-only LongFormer, suggesting that more data may be necessary to mitigate the discrepancy between pre-training of the language model and fine-tuning with early fusion (as in $SWST_{seq}$), in order to fully exploit the potential of the early fusion strategy. 
Our MMVTS Baseline$_{1,2,3}$ simply concatenate unimodal features to predict topic boundaries. Without pre-training on unlabeled data (PT, Eq.~\ref{formula_l_pre}) and the two auxiliary fine-tuning tasks to enhance multimodal coherence modeling (FT-Coh, Eq.~\ref{formula_l_finetune}), on F$_{1}$, MMVTS Baseline$_{1}$ outperforms the text-only Longformer by 2.28 and 2.90 on AVLecture and CLVTS respectively, while on \textit{Avg} score,  MMVTS Baseline$_{1}$ outperforms Longformer by 1.1 on AVLecture yet slightly underperforms Longformer by 0.65 on CLVTS. These results suggest that \textbf{\textit{simply concatenating unimodal features to predict topic boundaries does not bring consistent gains over unimodal models}}. 
The third group in Table~\ref{table_main_results} evaluates our MMVTS models with the four Multimodal Fusion Layer (MFL) architectures and with pre-training (PT) and fine-tuning (FT-Coh). \textbf{\textit{Overall, after PT and FT-Coh, on AVLecture, our MMVTS model using Co-Attention with MoE as MFLs outperforms all the competitive unsupervised and supervised visual-only and text-only baselines as well as the multimodal $SWST_{seq}$ and achieves the best \textit{Avg} (\textbf{4.52} absolute and \textbf{7.05\%} relative gain over previous SOTA), and the best\textit{BS@30} and \textit{F$_{1}$@30} results, setting new SOTA on AVLecture. On CLVTS, our MMVTS model achieves the best \textit{F$_{1}$} (5.21 absolute and 14.98\% relative gain over previous SOTA) and yields the \textit{Avg} score comparable to the best Avg score.}} On AVLecture, both gains on \textit{Avg} from our MMVTS model with Co-Attention MoE after PT and FT-Coh over the previous SOTA LongFormer$_{cssl}$ and MMVTS Baseline$_{1}$ are statistically significant ($p < 0.05$). Moreover, our MMVTS model significantly outperforms the multimodal model $SWST_{seq}$ on both datasets, demonstrating the effectiveness of our \textbf{\textit{middle fusion}} strategy and new pre-training and fine-tuning methods. Table~\ref{table_number_parameters} shows that our MMVTS model with Co-Attention with MoE has 192M trainable parameters while LongFormer$_{cssl}$ has 130M parameters. 
It is also notable that the performance of models on CLVTS is generally much lower than that on AVLecture, with the best \textit{Avg} on AVLecture and CLVTS differing by 17.50 (68.84 versus 51.34), indicating \textbf{\textit{a greater challenge to VTS from our CLVTS dataset than the AVLecture dataset}}.

\noindent \textbf{Comparison with fine-tuning text LLMs.} Table~\ref{table_main_results} also shows that on CLVTS, fine-tuning the powerful pre-trained text LLM Llama-3-8B with our \textit{Discrete} prompt (Appendix~\ref{appendix_llama}) achieves the best \textit{Avg} score 51.34, probably attributable to Llama-3-8B's extensive pre-training and vast knowledge base; still, the performance of our MMVTS model w/ Merge-Attn and MoE and Co-Attn and MoE after PT and FT-Coh is nearly the same, with Avg \textbf{51.33} and 51.27 respectively. However, on AVLecture, fine-tuning Llama-3-8B performs much worse than MMVTS model (60.26 versus our 68.61). Particularly, F$_{1}$ on both AVLecture and CLVTS from Llama-3-8B are much worse than MMVTS model as we find that Llama-3-8B's predicted boundaries often have a clip offset.  \textbf{\textit{These results underscore the value of our proposed small VTS models, since the efficiency and flexibility of our competitive to superior small models make them indispensable in many real-world applications}}. Future research could continue exploring how to integrate strengths of LLMs and multimodal approaches for VTS.

\noindent \textbf{Incorporate audio modality.} We also explore adding the audio modality (A) to the vision and text modalities (V+T) for our MMVTS model and find that V+T+A slightly improves the Avg score over V+T by 1.09\% and 3.32\% relatively, as shown in Appendix~\ref{appendix_audio}. Our ongoing research explores different audio features as well as directly employing visual and audio cues (i.e., V+A) for VTS.

\begin{table}[]
    \centering
    \resizebox{0.48\textwidth}{!}{
    \begin{tabular}{c|l|ccccc}
    \hline
        \textbf{PT} & \textbf{Model} & \textbf{\textit{F$_{1}$}} & \textbf{\textit{BS@30}} & \textbf{\textit{F$_{1}$@30}} & \textbf{\textit{mIoU}} & \textbf{\textit{Avg}}\\
    \hline
    \texttimes & Merge-Attn & 56.56 & 73.28 & 64.03 & 70.06 & 65.98$_{0.90}$\\
    \texttimes & Co-Attn & \textit{57.71} & 72.20 & 65.40 & 70.20 & 66.38$_{1.29}$\\
    \texttimes & Merge-Attn with MoE & 56.80 & 72.26 & 63.44 & 69.65 & 65.54$_{1.64}$\\
    \texttimes & Co-Attn with MoE & \textit{57.71} & \textit{74.30} & \textit{65.53} & \textit{71.45} & \textit{67.25$_{0.56}$}\\
    \hline
    \checkmark & Merge-Attn & 57.36 & 74.96 & 65.30 & 70.15 & 66.94$_{0.60}$\\
    \checkmark & Co-Attn & \textbf{60.01} & 73.88 & 67.27 & \textbf{72.32} & 68.37$_{0.52}$\\
    \checkmark & Merge-Attn with MoE & 57.54 & 73.48 & 64.36 & 70.43 & 66.45$_{0.14}$\\
    \checkmark & Co-Attn with MoE & 59.84 & \textbf{75.62} & \textbf{67.69} & 72.21 & \textbf{68.84$_{0.93}$}\\
    \hline
    \end{tabular}
    }
    \caption{Ablation studies of the pre-training tasks on AVLecture test set. The two auxiliary coherence modeling tasks are added in fine-tuning (Eq.~\ref{formula_l_finetune}). For \textbf{Avg}, we report mean and standard deviation from three runs with different random seeds.}
    \label{table_ablation_pretrain}
\end{table}

We conduct extensive ablation studies to validate effectiveness of the proposed Multimodal Fusion Layers, pre-training and fine-tuning tasks.

\noindent \textbf{(1) Effect of Multimodal Fusion Layers.} Comparing the third group and Baseline$_{3}$ in Table~\ref{table_main_results} shows that with the same pre-training and fine-tuning, the best performing architecture using Multimodal Fusion Layers always substantially outperforms Baseline$_{3}$. Specifically, with PT and FT-Coh, on AVLecture, both Co-Attention and Co-Attention with MoE notably outperform MMVTS Baseline$_{3}$ by 1.32 on \textit{Avg}; on CLVTS, all four Multimodal Fusion Layer architectures achieve remarkable gains on \textit{Avg} over MMVTS Baseline$_{3}$, from 2.65 to 4.18. These results demonstrate that \textbf{deep cross-modal interaction has notable advantage for multimodal fusion over simple unimodal feature concatenation for VTS}. Moreover, adding MoE on top consistently improves Co-Attention on both AVLecture and CLVTS, by 0.47 and 1.07 on \textit{Avg}; whereas, the effect of MoE on top of Merge-Attention is inconsistent, with a slight degradation on AVLecture and 1.53 gain on \textit{Avg} on CLVTS. In addition, we conduct more analysis of the effect of Co-Attn with MoE with different numbers of multimodal fusion layers in Appendix~\ref{appendix_mfl_number}.

\noindent \textbf{(2) Effect of Pre-training tasks.} Table~\ref{table_main_results} shows that for simple concatenation of unimodal features, pre-training before fine-tuning (as Baseline$_{3}$) outperforms Baseline$_{2}$ (w/o PT). We conduct ablation studies of the proposed pre-training and fine-tuning tasks on AVLecture. We apply the same fine-tuning with multimodal coherence modeling (FT-Coh, Eq.~\ref{formula_l_finetune}) and compare (a) random initialization of parameters for Multimodal Fusion Layers (w/o pre-training) (b) pre-training the model on unlabeled data. Table~\ref{table_ablation_pretrain} shows that w/o pre-training,  Co-Attn slightly improves \textit{Avg} over Merge-Attn by 0.4, and MoE further improves \textit{Avg} by 0.87. \textbf{\textit{Pre-training improves the performance on all four MFL architectures, with the average \textit{Avg} score increased by 1.36 (66.29~$\rightarrow$~67.65)}}. Pre-training also \textbf{\textit{improves model stability}} as the standard deviations of all w/ PT experiments are less than 1. Table~\ref{table_ablation_pretrain_task} in Appendix further compares the fine-tuning performance after applying different pre-training tasks. Compared to using both pre-training tasks, removing $l_{cma}$ or $l_{vts}$ degrades \textit{Avg} by 0.4 and 1.64 respectively,  indicating that the pretraining task aligned more closely with the downstream task, i.e., $l_{vts}$, yields greater gains.
\begin{table}[]
    \centering
    \resizebox{0.48\textwidth}{!}{
    \begin{tabular}{l|ccccc}
    \hline
        \textbf{Model} & \textbf{\textit{F$_{1}$}} & \textbf{\textit{BS@30}} & \textbf{\textit{F$_{1}$@30}} & \textbf{\textit{mIoU}} & \textbf{\textit{Avg}}\\
    \hline
    Co-Attn with MoE & 59.84 & \textbf{75.62} & \textbf{67.69} & 72.21 & \textbf{68.84}\\
    w/o $l_{cmal}$ & 58.96 & 74.62 & 67.39 & 72.17 & 68.29\\
    w/o $l_{mcssl}$ & 59.47 & 74.53 & 66.74 & \textbf{72.24} & 68.25\\
    w/o $l_{cmal}$ \& $l_{mcssl}$ & \textbf{60.57} & 73.36 & 66.52 & 70.42 & 67.72\\
    \hline
    \end{tabular}
    }
    \caption{Ablation studies of the two auxiliary coherence modeling fine-tuning tasks on AVLecture test set. Models are initialized from pre-training (Eq.~\ref{formula_l_pre}).}
    \label{table_ablation_coherence}
\end{table}

\noindent \textbf{(3) Effect of Fine-tuning tasks.} Table~\ref{table_ablation_coherence} compares different fine-tuning tasks after pre-training. Compared to the standard $l_{vts}$, adding the two auxiliary losses $l_{cma}$ and $l_{mcssl}$ notably improves  \textit{Avg} by 1.12, while decreasing $F_{1}$ by 0.73. 
These results suggest that \textbf{\textit{while multimodal coherence modeling may slightly compromise the precision of exact matches, it enhances the overall contextual comprehension of a model for VTS}}. Adding $l_{cma}$ or $l_{mcssl}$ individually improves \textit{Avg} by 0.53 and 0.57, with improvements mainly on \textit{BS@30} and \textit{mIoU}, suggesting that feature alignment at different granularities may improve fuzzy matching. 

\vspace{-3mm}
\section{Conclusion}
\vspace{-3mm}
We propose a novel supervised VTS model by thoroughly exploring multimodal fusion and coherence modeling. We also introduce a large-scale labeled Chinese Lecture dataset for VTS. Extensive experiments demonstrate the superiority of our model and effectiveness of critical algorithmic designs.

\section*{Limitations}
Our MMVTS model leverages the vision modal information encoded by the visual encoder. Considering the computational complexity, we keep the visual encoder frozen while keeping all other parameters trainable. This particular design may result in suboptimal utilization of the extensive multimodal information inherent in the dataset. We plan to further investigate different audio features in multimodal fusion (that is, V+T+A) as well as directly employing visual and audio cues (that is V+A since the current text modality is just ASR 1-best of the audio), to improve the VTS performance. We will continue enhancing the integration of general pre-trained multimodal models and large language models, which may offer a more holistic and effective exploration of multimodal information for VTS. Additionally, we will conduct more explorations and investigate approaches to make Co-Attn with MoE more stable in future work.


\bibliography{custom}

\appendix

\section{The CLVTS Dataset}
\label{subsec:dataset}

\begin{figure*}[htb]
    \centering
  \begin{minipage}[b]{0.32\linewidth}
    \centering
    \includegraphics[height=1.5in]{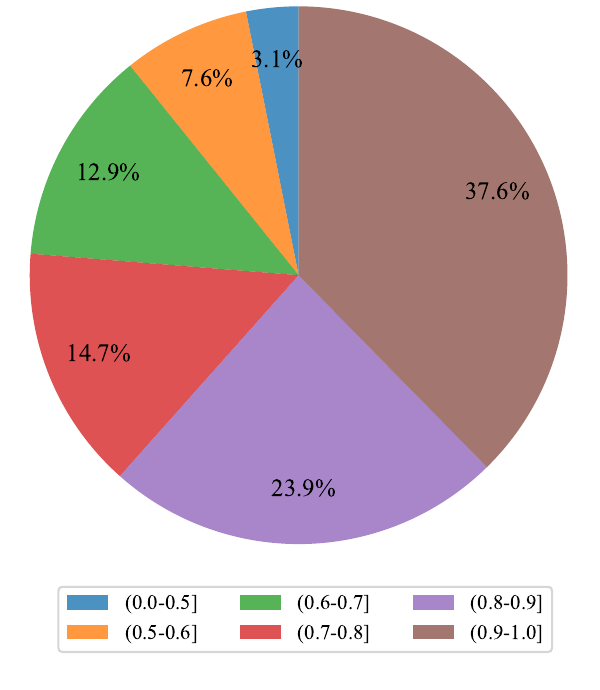}
    \subcaption{Distribution of Consistency Scores.}
    \label{figure_consistency_score}
  \end{minipage}
  \hfill
  \begin{minipage}[b]{0.32\linewidth}
    \centering
    \includegraphics[height=1.5in]{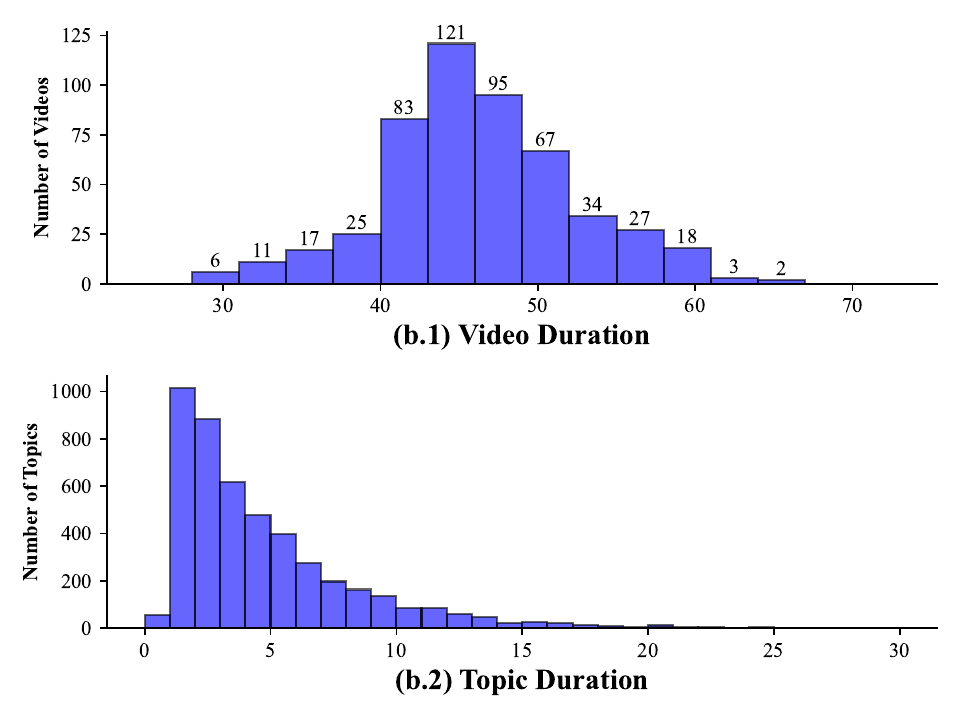}
    \subcaption{Video Duration and Topic Duration (in minutes).}
    \label{figure_video_topic_duration}
  \end{minipage}
  \hfill
  \begin{minipage}[b]{0.32\linewidth}
    \centering
    \includegraphics[height=1.5in]{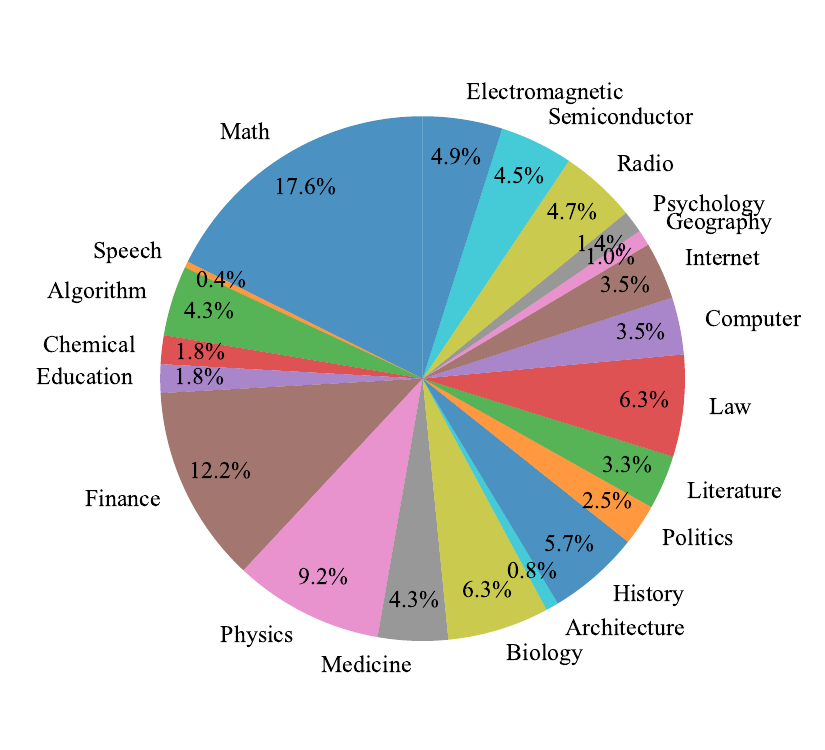}
    \subcaption{Distribution of Subjects of Lectures.}
    \label{figure_subject_distribution}
  \end{minipage}
  \caption{Statistics of our CLVTS dataset.}
  \label{figure_clvts_stats}
\end{figure*}

\subsection{\textbf{Data Collection and Annotation}}
The CLVTS dataset is primarily sourced from educational videos, in the form of lectures, from Public Video Platforms\footnote{\url{https://study.163.com/}}\footnote{\url{https://www.bilibili.com/v/knowledge}}. Specifically, videos are first transcribed by a competitive Automatic Speech Recognition (ASR) system\footnote{\url{https://tingwu.aliyun.com/home}}. We then ask annotators to combine visual and textual (ASR 1-best) information to mark the timestamp (in seconds) at the end of each topic. We ensure high accuracy and reliability of annotations from three aspects, including \textbf{annotator training}, \textbf{hierarchical topic labeling}, and \textbf{a multi-annotator strategy}. 

Firstly, before the actual annotation process starts, the annotators take two rounds of training. Each annotator needs to annotate 5 videos in each round; at the end of each round, we assess the annotation quality, provide feedback, and ensure that the annotators address the issues and understand the annotation guideline clearly, at the end of training. 

Secondly, during annotation, to help the annotators thoroughly understand the lecture content, we ask the annotators to annotate topics hierarchically, that is, they label both coarse-grained topic (large topic) and fine-grained topic (small topic) boundaries while the large topic boundaries are a subset of the small topic boundaries. We take the \textbf{small} topic boundaries as the final topic boundary labels for supervised VTS modeling.

Thirdly, we employ a multi-annotator strategy. All data is annotated in batches, with two annotators annotating each sample independently. The third annotator reviews the annotations by the first two annotators, rectifies errors, and provides the final annotations. After the annotations of each batch, we randomly select 5\% of a batch for quality assessment. If the unacceptable rate (the proportion of the wrongly annotated topic boundaries) is lower than 10\%, the data are deemed satisfactory and accepted; otherwise, after communicating quality assessment results and possible reasons for errors, the annotators are requested to re-annotate the batch based on the feedback. This quality control process is repeated until the unacceptable rate is lower than 10\% for all batches. In this work, we finished quality control of all data within 3 iterations.

\subsection{\textbf{Dataset Analysis}}
To evaluate the inter-annotator agreement on VTS annotations on the CLVTS dataset, following~\cite{shou2021generic}, we compute the $F_{1}@k$ score (defined in Appendix~\ref{appendix_metrics}) based on the absolute distance between two topic boundary sequences, varying the threshold $k$ from 0 to 8 seconds with a step size of 2 seconds, where 8 seconds are approximately the average duration of a video clip. By averaging the F$_{1}$ scores across all three pairs of annotated topic boundaries from three annotators on the same video, we obtain the \textbf{consistency score}. The more similar the annotations from all annotators on the same video are, the higher the consistency score is. Figure~\ref{figure_consistency_score} shows that the consistency scores of the majority of videos in our CLVTS dataset exceed 0.5, indicating a decent degree of consensus for VTS annotations~\cite{shou2021generic}.

Table~\ref{table_data_stats} compares our CLVTS dataset against existing VTS datasets. Both AVLecture and CLVTS are sourced from educational contexts, where VTS significantly enhances learning experiences. Table~\ref{table_data_stats} highlights that CLVTS features a higher average number of topics per video. Among annotated videos in CLVTS, 47\% are presentations showing slides, 34\% are blackboard demonstrations, and 19\% are miscellaneous types. \textbf{We also collect 1027 hours of unlabeled videos from the same sources for pre-training.} Figure~\ref{figure_video_topic_duration} and~\ref{figure_subject_distribution} show a diverse distribution of video durations and topic durations and a broad spectrum of subjects in labeled CLVTS.

\subsection{\textbf{Ethical Considerations}}
\label{appendix_ethical_considerations}
\textbf{The dataset used in this research is strictly for academic and non-commercial purposes.} We implemented several measures to ensure compliance with ethical standards, as follows.

\begin{itemize}
\item{\textbf{Data Transparency and Anonymization.}} We only provide ASR transcripts after rigorous text anonymization processes, visual features of video clips, our annotations, and links to the original videos, to ensures transparency regarding the data sources and their usage while maintaining anonymity.

\item{\textbf{Data Access Compliance.}} To further ensure ethical use of the dataset, we require researchers to contact us via emails to confirm their compliance with ethical guidelines and the conditions outlined in our data usage declaration, before granting them access to the dataset. This procedure includes ensuring that they are aware of and adhere to the Personal Information Protection Law (PIPL) and any relevant legal frameworks regarding personal data usage.

\item{\textbf{Authorization.}} Any personal data should be used only with express authorization, ensuring lawful and fair processing in accordance with applicable laws.
\end{itemize}

\begin{table}[]
    \centering
    \resizebox{0.32\textwidth}{!}{
    \begin{tabular}{l|c}
    \hline
    \textbf{Model} & \textbf{Numbers}\\
    \hline
        LongFormer$_{cssl}$ & 130M\\
        \hline
        \textbf{\textit{MMVTS Model (Ours)}} \\
        Baseline$_{1,2,3}$ & 154M\\
        Merge-Attn & 161M\\
        Co-Attn & 173M\\
        Merge-Attn with MoE & 175M\\
        Co-Attn with MoE & 192M\\
    \hline
    \end{tabular}
    }
    \caption{The number of trainable parameters of the baseline LongFormer$_{cssl}$ and variants of our MMVTS models. The model names conform to the model names in Table~\ref{table_main_results}.}
    \label{table_number_parameters}
\end{table}

\section{Implementation Details}
\label{appendix_training_details}
We partition the labeled data within AVLecture and CLVTS into 70\% for training, 10\% for validation, and 20\% for testing, respectively. The unlabeled data of AVLectures and CLVTS are used for pre-training for each dataset, respectively.

Our experiments are implemented with the \textit{transformers} package\footnote{\url{ https://github.com/huggingface/transformers}}. We use the same maximum sequence length 2048 as in~\citet{yu2023improving} for a fair comparison with the text-only models. \textbf{All results are the mean values over three runs with different random seeds.}

For video with the number of clips greater than the max sequence length, we use sliding window and take the last clip of the prior sample as the first clip of the next sample. All supervised models use a threshold strategy, where clips with scores above a threshold 0.5 are predicted as topic boundaries. 

Following~\cite{gupta2023unsupervised}, for each video clip, we extract three visual feature types: \textit{OCR}, \textit{2D} and \textit{3D}. Specifically, the \textit{OCR} features are derived by encoding the textual output obtained from the \textit{OCR} API\footnote{\url{https://help.aliyun.com/zh/viapi/developer-reference/api-sy75xq}} of the clip's central image. Encoding the textual output from OCR is performed using the BERT-based sentence transformer model\footnote{\url{https://www.sbert.net/docs/sentence_transformer/pretrained_models.html}}, where \textit{all-mpnet-base-v2} and \textit{paraphrase-multilingual-mpnet-base-v2} models are employed for experiments on the English and Chinese datasets, respectively. The \textit{3D} features are extracted using the same video feature extraction pipeline as in ~\citet{gupta2023unsupervised}. The \textit{2D} features are extracted by sampling three frames from each clip, subsequently encoding these frames with visual encoder, and applying max pooling. Specifically, we choose the visual encoder of CLIP~\cite{radford2021learning}, which is pre-trained to predict if an image and a text snippet are paired together. The images from AVLecture and CLVTS are processed to extract 2D features using CLIP$_{ViT-B/16}$\footnote{\url{https://github.com/openai/CLIP}} and CN-CLIP$_{ViT-B/16}$\footnote{\url{https://github.com/OFA-Sys/Chinese-CLIP}}, respectively.

\begin{figure}[tbp]
    \centering
    \includegraphics[width=0.5\textwidth]{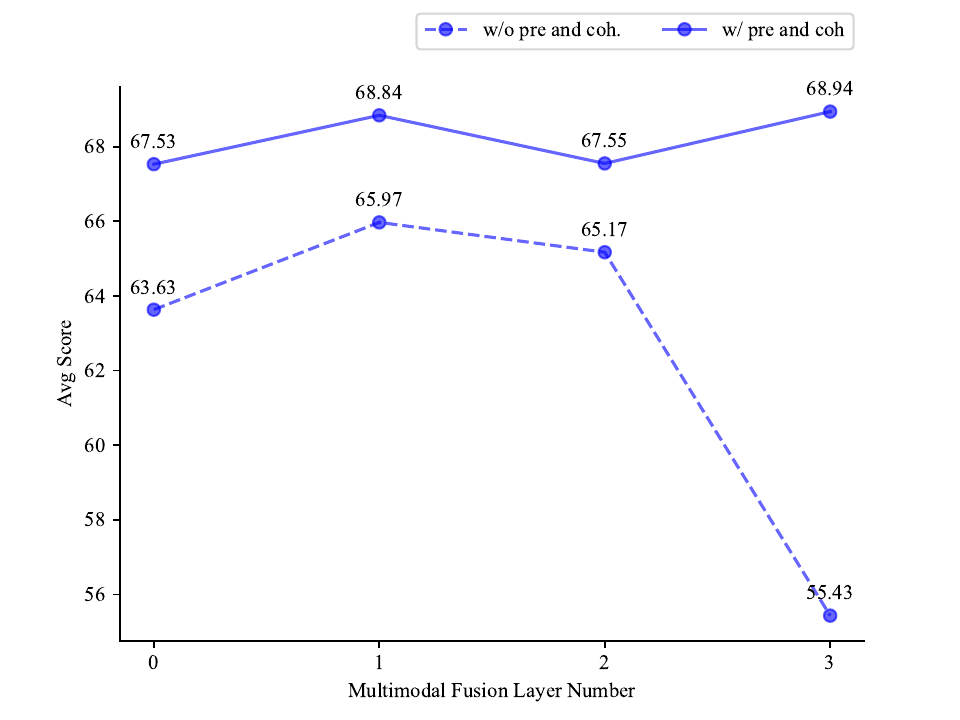}
    \caption{Performance (\textbf{Avg}) of fine-tuning our MMVTS model on AVLecture test set, w/o pre-training and multimodal coherence modeling tasks (denoted by w/o pre and coh) and w/ pre-training and multimodal coherence modeling tasks  (denoted by w/ pre and coh), with different numbers of Multimodal Fusion Layers, where the fusion structure is Co-Attention with MoE.}
    \label{figure_mfl_number}
\end{figure}

After extracting the visual features, we concatenate them as shown in Eq.~\ref{formula_cat_vis} to get $v_{i}$, which will then be fed into the following projection layer. During pre-training and fine-tuning, the parameters of visual encoders are kept frozen. The learning rate is $5e-5$ and dropout probability is 0.1. AdamW~\cite{loshchilov2017decoupled} is used for optimization. The batch size is 8 and the epoch for pre-training and fine-tuning is 1 and 5, respectively. The loss weight $\alpha$ and $\gamma$ for $l_{cma}$ is 0.5, $\beta$ and $\sigma$ for $l_{balance}$ is 1.0 when MoE is in Multimodal Fusion Layers, $\theta$ for $l_{mcssl}$ is 0.5. $k_{1}$ and $k_{2}$ of Eq.~\ref{formula_l_mcssl} are 1 and 3, following~\citet{yu2023improving}. We comprehensively compare different types of fusion structure using one Multimodal Fusion Layer, then use the best performing fusion structure for the remaining experiments. We also investigate the impact of different numbers of Multimodal Fusion Layers in Figure~\ref{figure_mfl_number}. As to the MoE layer in Multimodal Fusion Layers, we choose 4 candidate experts and activate 2 experts for each input feature. The intermediate size of expert is 3072. The total number of trainable parameters is shown in Table~\ref{table_number_parameters}.

\begin{equation}
    v_{i} = v^{2d}_{i};v^{3d}_{i};v^{ocr}_{i}
    \label{formula_cat_vis}
\end{equation}

\begin{table}[]
    \centering
    \resizebox{0.48\textwidth}{!}{
    \begin{tabular}{l|ccccc}
    \hline
        \textbf{Model} & \textbf{\textit{F$_{1}$}} & \textbf{\textit{BS@30}} & \textbf{\textit{F$_{1}$@30}} & \textbf{\textit{mIoU}} & \textbf{\textit{Avg}}\\
    \hline
    Co-Attn with MoE & 59.84 & \textbf{75.62} & 67.69 & \textbf{72.21} & \textbf{68.84}\\
    w/o $l_{vts}$ & 57.19 & 74.64 & 66.04 & 70.91 & 67.20\\    
    w/o $l_{cma}$ & \textbf{60.23} & 73.54 & \textbf{67.86} & 72.12 & 68.44\\
    \hline
    \end{tabular}
    }
    \caption{Ablation experiments of pre-training task on AVLecture. The model parameters derived from these distinct pre-training tasks served as the initial parameters for subsequent fine-tuning of the model. Additionally, the coherence modeling tasks are incorporated during the fine-tuning phase.}
    \label{table_ablation_pretrain_task}
\end{table}

\section{Evaluation Metrics}
\label{appendix_metrics}
\noindent \textbf{\textit{F$_{1}$}} is a metric used to evaluate the accuracy of text topic segmentation~\cite{lukasik2020text, zhang2021sequence}. It focuses on the performance of \textbf{exact matching} of the positive class and balances the precision and recall rates.

\noindent \textbf{\textit{BS@k}}~\cite{gupta2023unsupervised} is the average number of predicted boundaries matching with the ground truth boundaries within a k-second interval, which can be considered as the recall rate based on \textbf{fuzzy matching}.

\noindent \textbf{\textit{F$_{1}$@k}} denotes the \textit{F$_{1}$} score calculated based on matching predicted boundaries and ground truth boundaries within k seconds. Considering subjectivity and uncertainty in VTS annotations, we introduce \textit{F$_{1}$@k} as a supplement to \textit{BS@k} to enabling a more comprehensive assessment of model performance in dealing with ambiguous (uncertain) boundaries.

\noindent \textbf{\textit{mIoU}} is commonly used in video action segmentation~\cite{zhou2018towards} and video scene segmentation~\cite{mun2022bassl}. While the \textit{F$_{1}$}, \textit{BS@k} and \textit{F$_{1}$@k} metrics focus on the accuracy of positive predictions (either exact match or fuzzy match), the \textit{mIoU} metric measures the overlapping area between predicted segments and ground truth segments, hence providing a generalized assessment of how well the model's predicted segments match the actual segments on the segmentation task.

Our implementation of \textit{BS@30} draws upon the code published by~\citet{gupta2023unsupervised}\footnote{\url{https://github.com/Darshansingh11/AVLectures/}}, while the approach to implement \textit{mIoU} is guided by~\citet{mun2022bassl}\footnote{\url{https://github.com/kakaobrain/bassl}}. We have relied on the \textit{scikit-learn} package\footnote{\url{https://scikit-learn.org/stable/}} to compute  \textit{F$_{1}$}, following the implementation by~\citet{yu2023improving}, to ensure fair comparisons. The definitions provided in the aforementioned sources also inform our implementation of \textit{F$_{1}$@k}.

\section{Artifact Use Consistent With Intended Use}

All of our use of existing artifacts is consistent with their intended use, provided that it was specified. For the CLVTS data set we created, its license will be for research purpose only.

\begin{figure*}[t]
    \centering
    \begin{subfigure}{0.48\textwidth}
        \includegraphics[width=\linewidth]{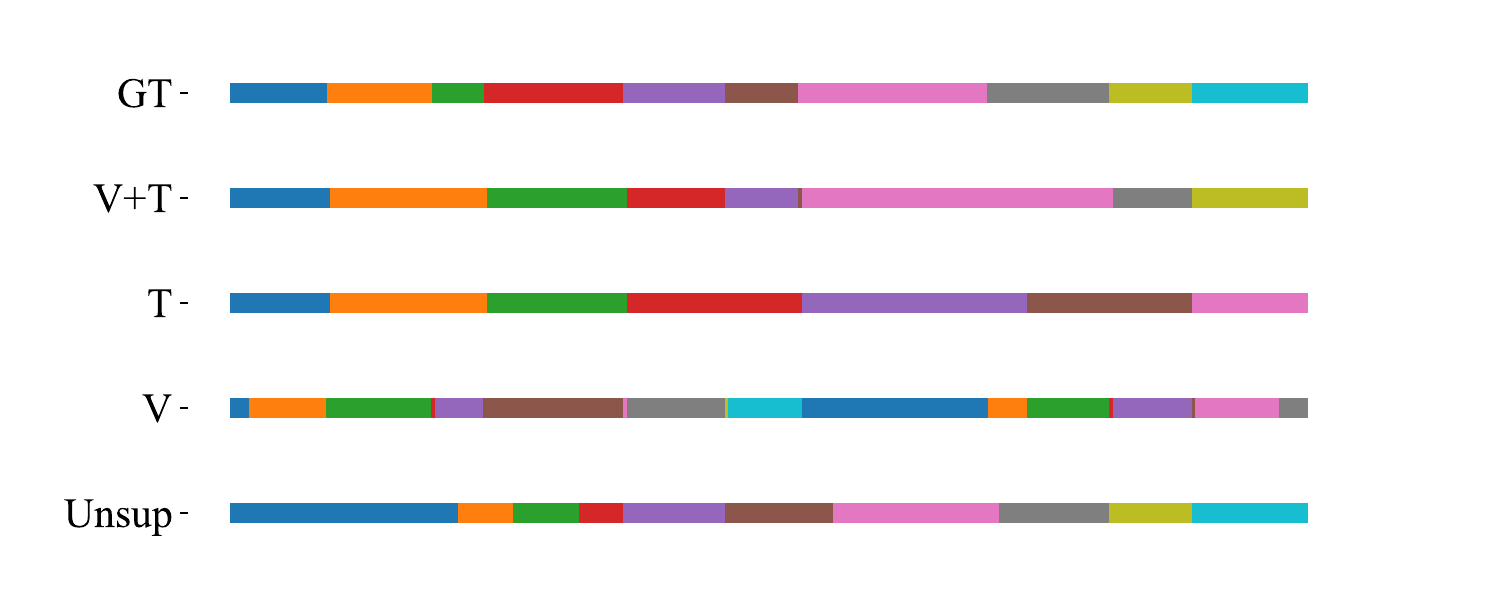}
        \caption{Introduction to Nuclear and Particle Physics}
        \label{fig:image1}
    \end{subfigure}
    \begin{subfigure}{0.48\textwidth}
        \includegraphics[width=\linewidth]{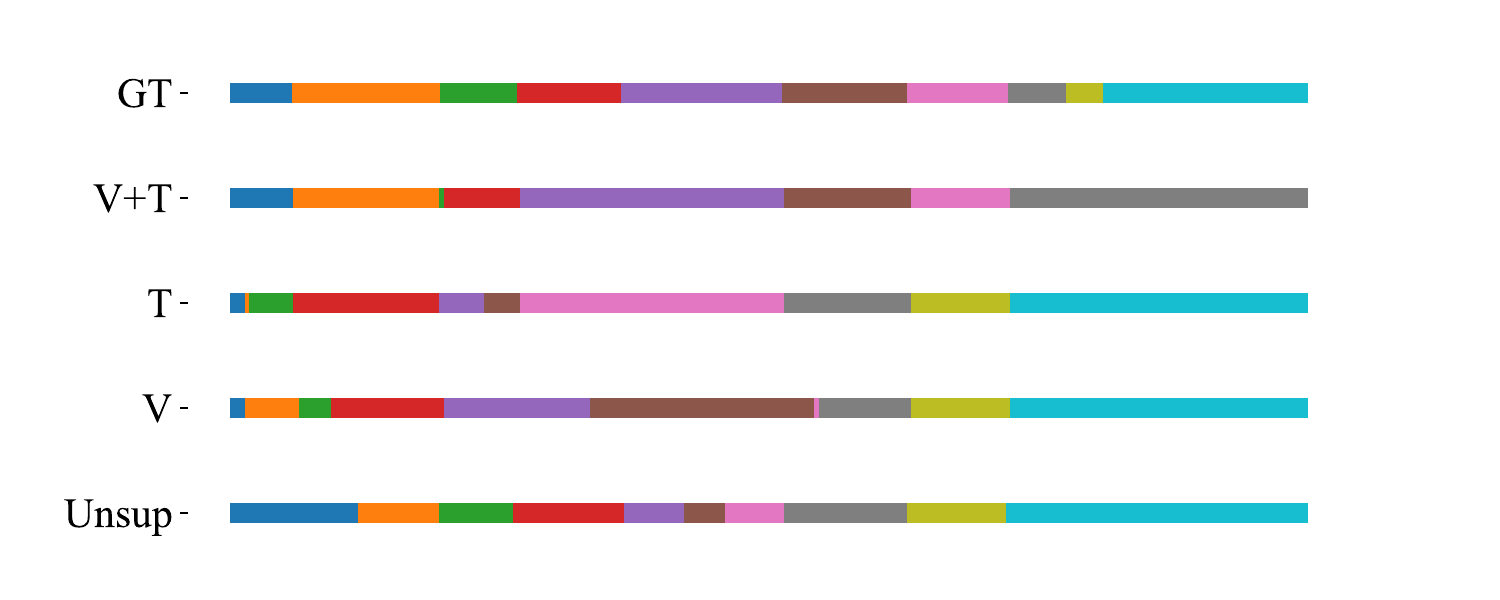}
        \caption{Latency and Throughput}
        \label{fig:image2}
    \end{subfigure}
    
    \begin{subfigure}{0.48\textwidth}
        \includegraphics[width=\linewidth]{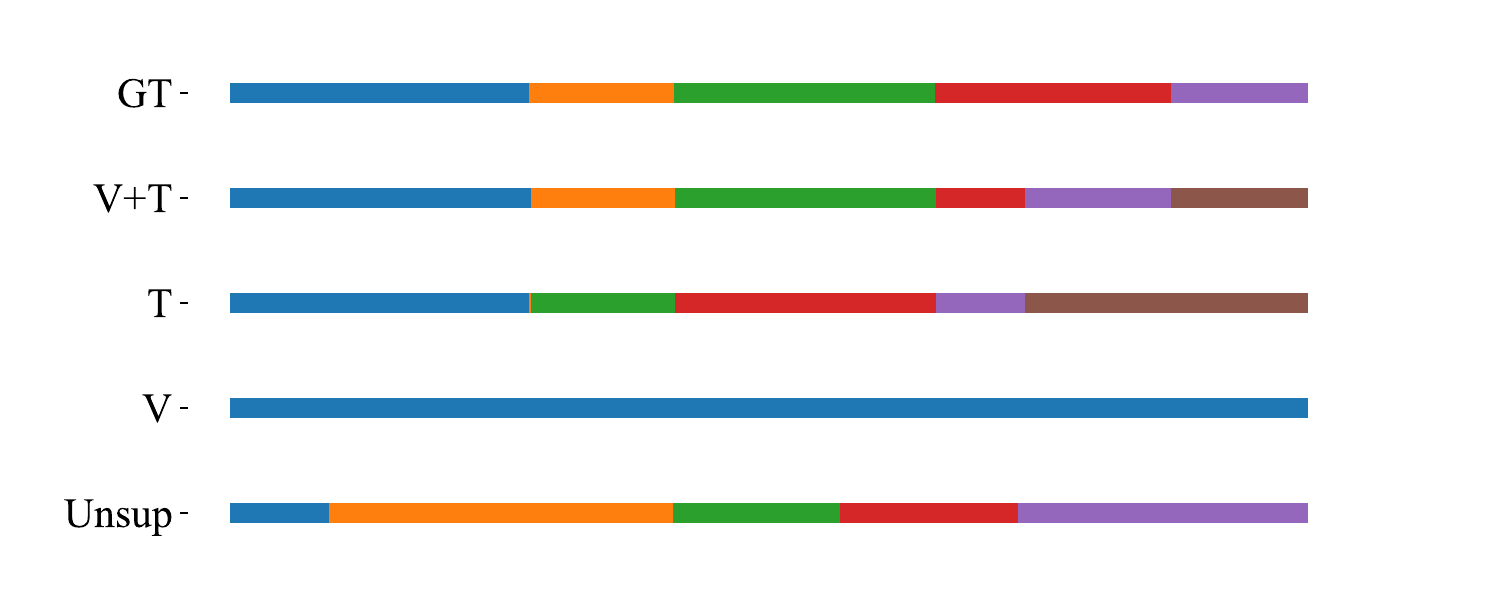}
        \caption{Quiz Review From Optional Problem Set 8}
        \label{fig:image3}
    \end{subfigure}
    \begin{subfigure}{0.48\textwidth}
        \includegraphics[width=\linewidth]{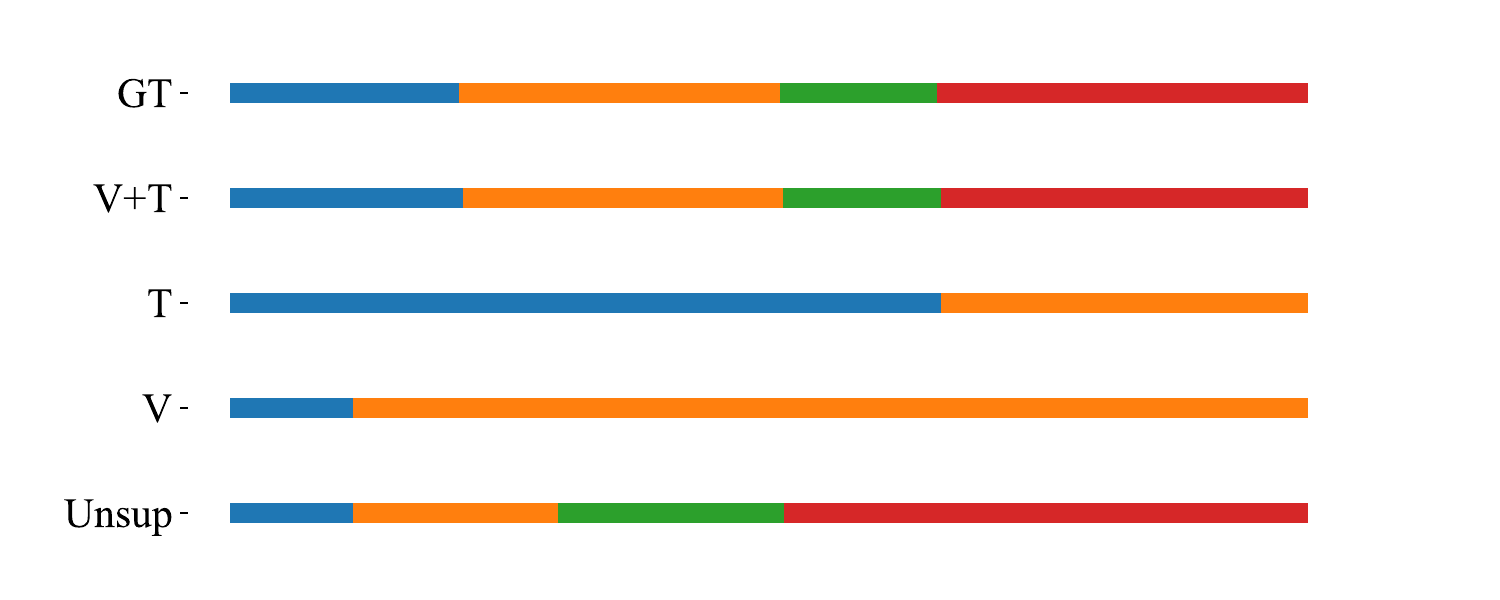}
        \caption{Quantum Mechanics}
        \label{fig:image4}
    \end{subfigure}
    
    \begin{subfigure}{0.48\textwidth}
        \includegraphics[width=\linewidth]{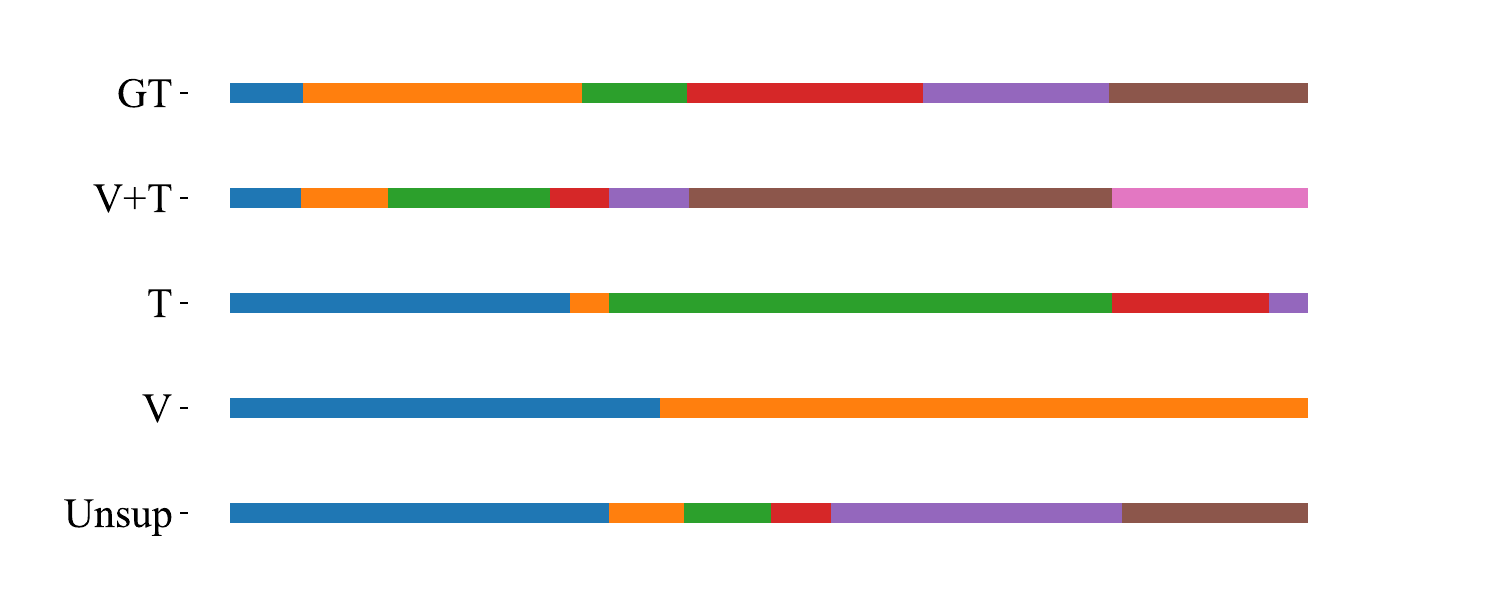}
        \caption{Attention}
        \label{fig:image5}
    \end{subfigure}
    \begin{subfigure}{0.48\textwidth}
        \includegraphics[width=\linewidth]{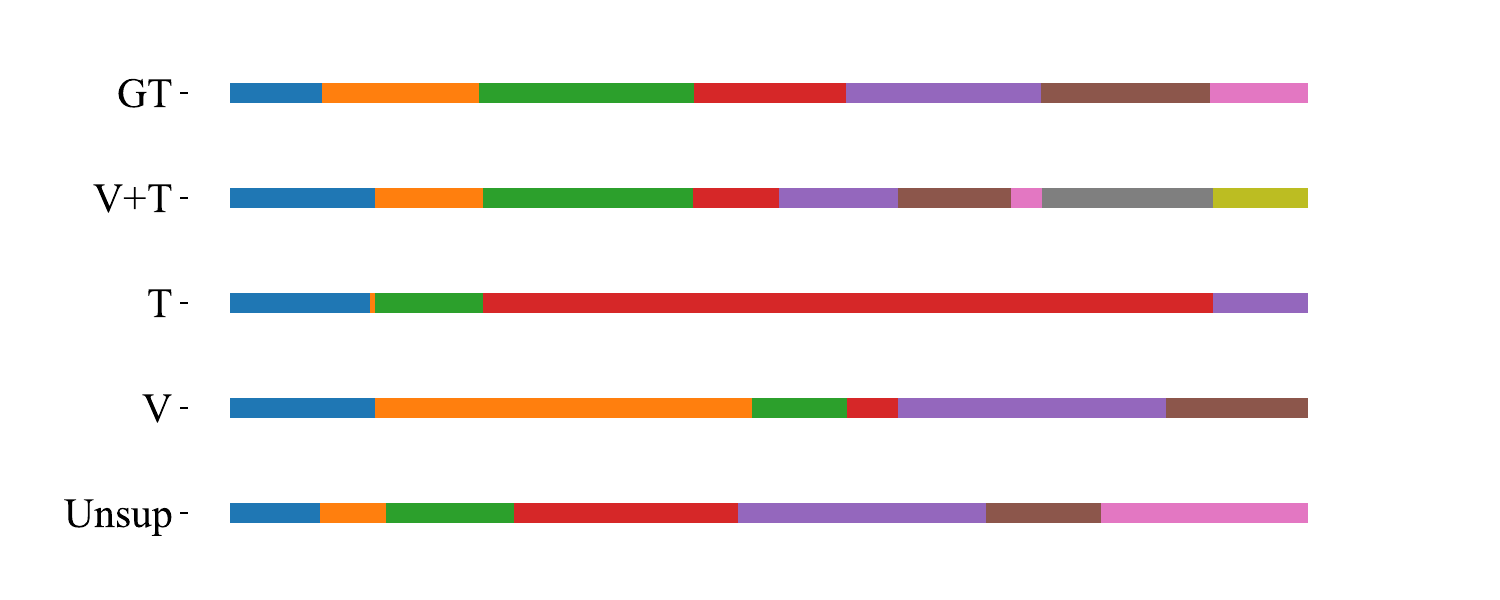}
        \caption{Vision}
        \label{fig:image6}
    \end{subfigure}
    
    \caption{Video Topic Segmentation examples for six lecture videos from AVLecture. (a)-(b) are of the \textbf{Slides} mode, (c)-(d) are \textbf{Blackboard} mode and (e)-(f) are \textbf{Mixed} mode. \textbf{GT} denotes ground truth, \textbf{V} refers to the visual-only baseline BaSSL which only uses visual features, \textbf{T} denotes the text-only LongFormer which only relies on the text modality, and \textbf{V+T} signifies our MMVTS model that integrates information from both visual and text modalities. \textbf{Unsup} denotes the unsupervised baseline UnsupAVLS, which also integrates visual and text features.}
    \label{fig_qualitative_analysis}
\end{figure*}

\section{Fine-tune Llama-3-8B for Video Topic Segmentation}
\label{appendix_llama}
Considering the computational complexity and the data volume, we employ LoRA~\cite{hu2021lora} for fine-tuning Llama-3-8B\footnote{\url{https://huggingface.co/meta-llama/Meta-Llama-3-8B}} on the training set of each AVLecture and CLVTS datasets, with a maximum sequence length of 2048. With LoRA, the number of trainable parameters is 3 million. Our training configuration includes a batch size of 32 and a total of 6 epochs, utilizing a learning rate of $5e-5$ and \textit{cosine} scheduler~\cite{loshchilov2016sgdr}. Table~\ref{table_prompt_for_ts} shows two prompts we used to fine-tuning Llama-3-8B on the text modality of VTS data. We first evaluate the \textit{Generative} type prompt that~\citet{yu2023improving} has experimented with, due to its better zero-shot and one-shot text topic segmentation performance than the \textit{Discriminative} type. However, compared with Longformer which also uses the textual modality, the \textit{Avg} score of fine-tuning Llama-3-8B with the \textit{Generative} prompting strategy (denoted by Llama-3-8B$_{Generative}$) is 8.3 and 6.27 absolutely worse than those from Longformer on AVLecture and CLVTS, respectively. We suspect that this is due to the inherent issue of sparse labels in binary classification for VTS, posing challenges to applying the large language model (LLM) Llama-3-8B in a generative manner. Inspired by this hypothesis, we refine the \textit{Discriminative} prompt into the \textit{Discrete} prompt for Llama-3-8B, as shown in Table~\ref{table_prompt_for_ts}. The results in Table~\ref{table_main_results} show that on \textbf{Avg}, Llama-3-8B$_{Discrete}$ consistently and markedly outperforms Llama-3-8B$_{Generative}$ on both AVLecture and CLVTS, by 6.04 and 10.8 absolute points, respectively. These results suggests that the \textit{Discrete} prompt is more suitable for prompting Llama-3-8B for the VTS task than the Generative prompt. 

However, with either Generative or Discrete prompt, Llama-3-8B$_{Discrete}$ does not provide consistent advantages over the text-only small model Longformer. Table~\ref{table_main_results} shows that $F_{1}$ results of Llama-3-8B$_{Discrete}$ on AVLecture and CLVTS are much worse than those from Longformer (we find that the prediction boundary of Llama-3-8B usually has a clip offset). The \textit{Avg} score of Llama-3-8B$_{Discrete}$ on AVLecture is 2.26 absolute worse than LongFormer, although Llama-3-8B$_{Discrete}$ achieves the best \textit{Avg} score on CLVTS, surpassing LongFormer by 4.53 absolute.  Future work could further explore how to better use LLMs for the VTS task.

\begin{table}[tbp!]
    \centering
    \resizebox{0.5\textwidth}{!}{
        \begin{tabular}{c|p{10cm}}
        \hline
            \textbf{Type} & \textbf{Prompts for text topic segmentation.} \\
            \hline
            \multirow{7}*{\textit{Generative}} & Please identify several topic boundaries for the following document and each topic consists of several consecutive utterances. please output in the form of \{topic i:[], ... ,topic j:[]\} with json format, where the elements in the list are the index of the consecutive utterances within the topic, and output even if there is only one topic.\\
            & document: \\
            & [0]: $s_{1}$ \\
            & [1]: $s_{2}$ \\
            & ... \\
            & [$n-1$]: $s_{n}$ \\
            & Please give the result directly in json format:\\
            & output: \{"topic\_0": [0, 1, 2, …, k-1], "topic\_1": [k, k+1, …], …\}\\
            \hline
            \multirow{8}*{\textit{Discrete}} & Please identify several topic boundaries for the following document. please output in the form of \{topic\_segment\_ids:[xxx]\} with json format, where the elements in the list are the index of the last sentence of every topic, if there is only one topic then the array is empty.\\
            & document:\\
            & [0]: $s_{1}$ \\
            & [1]: $s_{2}$ \\
            & ... \\
            & [$n-1$]: $s_{n}$ \\
            & Please give the result directly in json format:\\
            & output: \{"topic\_segment\_ids": [x, x, x]\}\\
            \hline
        \end{tabular}
    }
    \caption{Our designed prompts for fine-tuning Llama-3-8B on the text data for video topic segmentation. $n$ denotes the number of sentences and $s_{i}$ denotes the $i$-th sentence in the document.}
    \label{table_prompt_for_ts}
\end{table}

\section{More Analysis of the Baseline Results}
\label{appendix_more_analysis_baseline_results}
Considering the baseline results in Table~\ref{table_main_results}, the only seemingly similar scores between the unsupervised method and supervised baselines are mIoU scores, which are attributed to the leakage of the ground-truth topic number. If we use the topK probability to determine predictions, where K for each sample is the ground-truth topic number, mIoU and BS@30 of $SWST_{seq}$  (Row 7) will be 4.4 and 19.05 points higher than UnsupAVLS, respectively. However, this evaluation is not reasonable, as models should not disclose the ground-truth topic number during testing. Therefore, we opted for the threshold-based evaluation strategy commonly used in classification tasks

\section{Performance of Multimodal Fusion Layers with Varied Numbers of Layers}
\label{appendix_mfl_number}
Using Co-Attention with MoE as the architecture of Multimodal Fusion Layers, we investigate the \textit{Avg} performance of models featuring various numbers of Multimodal Fusion Layers (MFLs) on the AVLecture data set, as depicted in Figure~\ref{figure_mfl_number}. We find that directly fine-tuning our MMVTS model, without pre-training nor the two auxiliary tasks for coherence modeling, a single MFL yields the best performance, surpassing the no-layer configuration by 2.34 points. Adding more layers leads to degraded performance and training instability, particularly noticeable with three layers. 

By incorporating the pre-training phase followed by fine-tuning with coherence modeling tasks, Avg performance enhancements of 4.0, 2.87, 2.37, and 13.51 points for 0, 1, 2, 3 MFL layers are observed, respectively. These results clearly demonstrate the substantial benefits from our pre-training and coherence modeling strategies in boosting the model's performance. Notably, our pre-training and coherence modeling reaches the convergence of the model with three MFL layers, achieving results that marginally exceed the performance of a single-layer model by 0.1 points.

\begin{table}[]
    \centering
    \resizebox{0.45\textwidth}{!}{
    \begin{tabular}{l|l|c|c}
    \hline
     \textbf{Model} & \textbf{Modality} & \textbf{AVLecture} & \textbf{CLVTS} \\
    \hline
    Longformer & T & 62.52 & 46.81 \\
    Llama-3-8B$_{Discrete}$ & T & 60.26 & 51.34 \\
    \hline
    \multirow{3}{*}{MMVTS Model (Ours)} & V+T & 68.61 & 51.27 \\
    & V+T+A$_{1}$ & 67.49 & 51.98 \\
    & V+T+A$_{2}$ & \textbf{69.36} & \textbf{52.97}\\
    \hline
    \end{tabular}
    }
    \caption{The \textbf{Avg} score of integration of audio information into the MMVTS model with Co-Attn and MoE architecture, on AVLecture and CLVTS test sets. V+T+A$_{1}$ notes that during fine-tuning phase, the audio features are incorporated into the cross-modality alignment loss l$_{cma}$, while V+T+A$_{2}$ does not involve audio features in l$_{cma}$.}
    \label{table_audio}
\end{table}

\section{Qualitative Analysis of Video Topic Segmentation Examples}
\label{appendix_qualitative_analysis}
We present the topic-segmented outputs for six lecture videos from three presentation modes including \textbf{slides}, \textbf{blackboard}, and \textbf{mixed}, in Figure~\ref{fig_qualitative_analysis}. Relying solely on the visual modality, segmentation points are predominantly identified through the superficial cues associated with visual transitions. In contrast, using only the textual modality, topic boundaries are discerned based on semantic content; however, this approach has limitations, such as missing topic boundaries or the accumulation of topics. The integration of the visual modality with the text modality offers complementarity of information, thereby improving the overall VTS performance. The case studies in Figure~\ref{fig_qualitative_analysis} help illustrate the importance of multimodal fusion for the VTS task regardless of the video presentation modes.

\section{Integration of Audio Information}
\label{appendix_audio}

We use a pre-trained Whisper-small model\footnote{\url{https://github.com/openai/whisper/}} with 244 million parameters and use the Whisper encoder to extract audio features for video clips (the parameters of the Whisper audio encoder are frozen). Similar to visual features, we perform maximum pooling on the encoded audio features for each clip to obtain the clip-level audio features. Then, we feed the audio features into a learnable projection layer and then through the Multimodal Fusion Layers, and then concatenate them with visual features and text features for the final topic boundary prediction. 

We initialize the parameters from pre-trained visual and textual MMVTS model with Co-Attn and MoE. During fine-tuning, we compare two configurations. The first configuration is denoted by V+T+A$_{1}$, which incorporates the audio feature into the cross-modality alignment loss l$_{cma}$. The second configuration is denoted by V+T+A$_{2}$, which does not involve audio features in l$_{cma}$. The pairwise cross-modality alignment loss weight is set to 0.33 in V+T+A$_{1}$.

The results are shown in Table~\ref{table_audio}. As can be seen from the table, integration of audio features by V+T+A$_{1}$ improves the \textit{Avg} score by 0.71 on the CLVTS test set while decreasing the \textit{Avg} socre by 1.12 on the AVLecture test set. However, we find that excluding audio from the l$_{cma}$ (V+T+A$_{2}$) results in an absolute improvement of 0.75 and 1.7 in the \textit{Avg} score on AVLecture and CLVTS test sets, respectively, which suggests that audio features can contribute to more accurate topic boundary prediction under certain conditions, but their role in modality alignment needs to be treated with caution and demands further exploration. We sample some videos and observe that the cosine similarity between audio features of adjacent clips shows relatively small differences, ranging from 0.97 to 0.98, while visual features between adjacent clips exhibit larger differences, ranging from 0.88 to 0.97. This disparity might complicate the task of simultaneously aligning text, visual, and pooled clip-level audio features. Future research could explore more nuanced integration of audio features to provide supplementary paralinguistic information, such as pitch, energy, and pause duration. Alternatively, direct use of audio and visual information for VTS may bypass the ASR step altogether (as ASR is used to obtain the text modality).

\end{document}